\documentclass{article} 
\usepackage{iclr2027_conference,times}

\usepackage{amsmath,amsfonts,bm}

\def\eqref#1{equation~\ref{#1}}

\def\1{\bm{1}}

\DeclareMathAlphabet{\mathsfit}{\encodingdefault}{\sfdefault}{m}{sl}
\SetMathAlphabet{\mathsfit}{bold}{\encodingdefault}{\sfdefault}{bx}{n}

\usepackage{hyperref}
\usepackage{url}
\usepackage{booktabs}
\usepackage{graphicx}
\usepackage{amssymb}
\usepackage{xspace}
\usepackage{xcolor}
\usepackage{colortbl}
\usepackage{multirow}
\usepackage{pifont}
\usepackage{flafter}
\usepackage{array}
\usepackage{placeins}
\usepackage{sections/prompt_boxes}

\definecolor{scoregreen}{HTML}{246B8E}
\definecolor{scorered}{HTML}{A45337}
\definecolor{scoregray}{gray}{0.30}
\definecolor{paperblue}{HTML}{246B8E}
\definecolor{paperteal}{HTML}{36877D}
\definecolor{paperink}{HTML}{263746}
\hypersetup{
  colorlinks=true,
  citecolor=paperblue,
  linkcolor=paperblue,
  urlcolor=paperblue,
  pdftitle={Learning from Others, Acting for You: Cross-User Memory Sharing for LLM Agents}
}

\newcommand{\cmark}{\ding{51}}

\usepackage{tikz}
\usetikzlibrary{arrows.meta,positioning,shapes.geometric,fit,calc,backgrounds}

\usepackage{xspace}

\newcommand{\methodname}{ShareMem\xspace}

\title{Learning from Others, Acting for You: Cross-User Memory Sharing for LLM Agents}

\iclrfinalcopy

\author{
Jinming Hu\textsuperscript{\rm 1}\thanks{These authors contributed equally.} , Haodong Zhao\textsuperscript{\rm 1}\textsuperscript{,}\textsuperscript{\rm 3}\footnotemark[1] , Qi Jia\textsuperscript{\rm 2}\thanks{Corresponding authors.} , \textbf{Die Chen}\textsuperscript{\rm 3}, \textbf{Tianhang Zhao}\textsuperscript{\rm 1},\\ \textbf{Sufeng Duan}\textsuperscript{\rm 1}, \textbf{Gongshen Liu}\textsuperscript{\rm 1}\footnotemark[2]\\
$^{1}$School of Computer Science, Shanghai Jiao Tong University \\
$^{2}$Shanghai Artificial Intelligence Laboratory \\
$^{3}$School of Computing, National University of Singapore \\
\texttt{\{hujinming, zhaohaodong\}@sjtu.edu.cn},
\texttt{jiaqi@pjlab.org.cn}, \\ 
\texttt{daphnechen@u.nus.edu}, 
\texttt{\{zthzthzth, 1140339019dsf, lgshen\}@sjtu.edu.cn}
}

\begin{document}

\maketitle

\begin{abstract}
Large language model (LLM) agents serving different users often solve related tasks, yet separate user histories can leave reusable experience inaccessible to other agents. Pooling memories expands access but risks transferring preferences that conflict with the receiving user's requirements. We introduce \methodname, a memory architecture that shares reusable experience while grounding its application in the receiving user's own preferences. Shared experiences indicate how to act and which preferences to consult; the receiving user's memory supplies their concrete values. Two-stage consolidation refines experience locally before integrating accepted edits into a shared pool. During execution, scope-first retrieval jointly selects local and shared experiences under a common entry budget, while a user-bound channel supports initial and agent-initiated preference retrieval. We evaluate \methodname across web navigation (Mind2Web), online personalized interaction (VitaBench~2.0), and multi-session coding (MemoryCode) with four backbone models. It improves step success, average task success, and dialogue-macro coding scores, respectively, over matched user-local memory across all four models. Ablations favor two-stage consolidation for smaller shared pools, lower induction token usage, and better downstream performance, and support complementarity between experience guidance and active preference retrieval. Further analyses show that sharing helps most when relevant local experience is scarce, while source quality and cross-user preference interference limit useful transfer.

\end{abstract}

\section{Introduction}
\label{sec:introduction}

Long-term memory enables large language model (LLM) agents to reuse past experience and personalize assistance~\citep{park2023generative,packer2023memgpt,zhong2024memorybank,chhikara2025mem0}. Consider an organization where employees use the same agent system, each with a personalized instance and a separate interaction history. These histories contain reusable procedures for common tasks alongside personal preferences and requirements. Experience acquired by one instance may therefore benefit others, provided that its application respects the receiving user's constraints.

Keeping these histories local preserves their association with individual users, but also isolates useful experience: an agent may lack guidance already acquired by another for a similar task. We call this an \emph{experience-reachability bottleneck}. Pooling histories expands access, but can introduce instructions that conflict with the receiving user's requirements. For example, a coding procedure may be reusable across employees, while the contributor's preferred naming convention may not be.

Existing works distill interaction histories into lessons, skills, or workflows for subsequent tasks~\citep{shinn2023reflexion,zhao2024expel,wang2024agent}. Shared repositories, collaborative memory, and federated skill learning extend reuse across agents through experience aggregation, access control, and collective refinement~\citep{gao2024memory,rezazadeh2025collaborative,ma2026skillclaw,yang2026federatedskill}. However, making experience accessible does not establish its applicability to another user. Access permissions govern retrieval, while successful personalization also requires separating reusable procedures from user-specific preferences. This motivates our central question: \emph{How can agents reuse others' experience while applying their own user preferences?}

We introduce \methodname, a memory architecture that shares guidance about how to perform tasks and which personal constraints to consult, while resolving concrete values through the receiving user's own memory. Each agent maintains local experiences and personal preferences. Two-stage consolidation extracts reusable local experience and consolidates accepted edits in a shared pool. During execution, scope-first retrieval jointly selects local and shared experiences under a joint entry budget, while a user-bound preference channel supports initial and agent-initiated recall of personal constraints. Across web navigation, online personalized interaction, and multi-session coding with four backbone models, comparisons with matched user-local configurations show gains from sharing, especially where relevant local experience is sparse. Further analyses examine consolidation and active preference retrieval, and identify source quality and cross-user preference interference as limits to useful transfer. Our contributions are:

\textbf{Experience-reachability perspective.} We identify the isolation of reusable experience as a bottleneck for personalized agents, distinguishing access to useful procedures from the applicability of user-specific values.

\textbf{Selective memory sharing architecture.} We introduce \methodname, which combines two-stage experience consolidation and joint local--shared retrieval with a separate preference channel bound to the receiving user.

\textbf{Empirical characterization of transfer.} We characterize when sharing helps and where it falls short, examining local coverage, memory consolidation, retrieval behavior, and interference from other users' preferences.

\section{Related Work}
\label{sec:related-work}

\paragraph{Individual agent memory and personalization.}
Generative Agents integrates experience streams with reflection and retrieval~\citep{park2023generative}, while CoALA places memory within agent cognition~\citep{sumers2023cognitive}. MemGPT, MemoryBank, Mem0, and A-Mem develop long-term storage and retrieval~\citep{packer2023memgpt,zhong2024memorybank,chhikara2025mem0,xu2026mem}, and personalized memory systems adapt assistance through user-grounded retrieval and interaction histories~\citep{salemi2024lamp,wang2024crafting,tan2025prospect}. Building on this user-specific grounding, we study how agents can also benefit from experience outside their own histories.

\paragraph{Abstracting and maintaining experience.}
Beyond retaining histories, experience abstraction produces reusable guidance. Reflexion and ExpeL extract feedback and lessons~\citep{shinn2023reflexion,zhao2024expel}, while Voyager and Agent Workflow Memory develop reusable skills and workflows~\citep{wang2023voyager,wang2024agent}. ReasoningBank, Agentic Context Engineering, and MCMA explore abstraction, refinement, and maintenance~\citep{ouyang2026reasoningbank,zhang2026agentic,liang2026learning}. When experience is shared across users, redundant contributions can compete for limited retrieval slots. \methodname addresses this through two-stage consolidation: refining local experience before integrating accepted edits into a shared pool.

\paragraph{Sharing experience across users and agents.}
Memory Sharing aggregates agent experience~\citep{gao2024memory}, while Collaborative Memory manages multi-user sharing under asymmetric access control~\citep{rezazadeh2025collaborative}. SkillClaw, FederatedSkill, and lifelong multi-agent memory systems study collective skill evolution and experience reuse~\citep{ma2026skillclaw,yang2026federatedskill,wu2026scaling}. Agent KB and FedWorld examine knowledge applicability across environments or clients~\citep{tang2025agent,hou2026fedworld}. Retrieval permission and contextual relevance alone do not establish whether a preference applies to the receiving user. \methodname therefore pairs scope-first experience retrieval with initial and active recall from that user's own memory, separating transferable guidance from personal constraints. Our interference analysis examines this boundary.

\section{\methodname: Selective Memory Sharing}
\label{sec:method}

\begin{figure}[h]
    \centering
    \includegraphics[width=\textwidth]{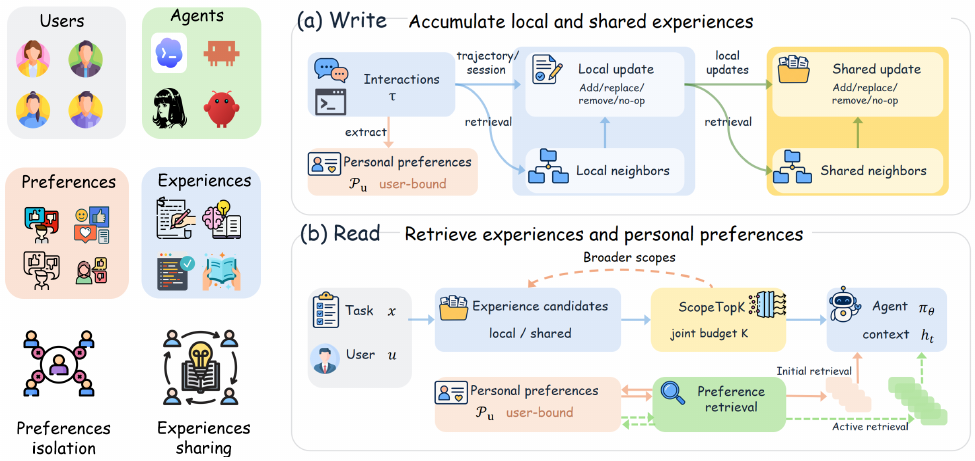}
\caption{\textbf{\methodname overview.} Left: agents share reusable experiences while keeping preferences user-bound. (a) Two-stage consolidation: interactions and retrieved local experiences inform local updates; accepted edits then provide evidence for consolidation with shared experiences. (b) Scope-first retrieval selects local and shared experiences under a joint budget $K$. A separate channel supports initial and agent-initiated retrieval of the receiving user's preferences.}
\label{fig:method}
\label{fig:formation}
\label{fig:retrieval}
\end{figure}

\methodname enables agents to reuse experience across users while grounding its application in the receiving user's preferences. Figure~\ref{fig:method} summarizes two-stage experience consolidation, scope-first local--shared retrieval, and a separate preference channel supporting initial and active recall.

\subsection{Memory Structure}
\label{sec:design}

We consider a multi-user deployment, such as an organization or a shared agent platform, where each user $u$ has a dedicated agent instance $A_u$ and interaction history $\mathcal H_u$. Users may face similar tasks while differing in their preferences and requirements.

Local memory $\mathcal M_u$ comprises preferences $\mathcal P_u=\{p_u^1,\ldots,p_u^{n_u}\}$ and experiences $\mathcal E_{u,s}=\{e_{u,s}^1,\ldots,e_{u,s}^{m_{u,s}}\}$ for applicability scope $s$. Here $n_u$ and $m_{u,s}$ are current collection sizes. The $i$th preference $p_u^i$ records a personal requirement; the $i$th experience $e_{u,s}^i$ describes a reusable procedure and relevant preference dimensions. Shared memory $\mathcal S$ comprises cross-user experience collections $\mathcal R_s$ under the same scopes. An environment-specific adapter assigns scopes and retrieval priorities.

Preferences are maintained from each user's history separately from experience consolidation (Appendix~\ref{app:retrieval-policy}). During execution, agents retrieve procedural guidance from local and shared experiences and resolve personal requirements through the receiving user's $\mathcal P_u$. Thus, sharing expands experience access without granting access to other users' preference stores.

\subsection{Two-stage experience consolidation}
\label{sec:experience-write}

For a trajectory or session $\tau\in\mathcal H_u$ eligible under the evaluation protocol, let $s=\sigma(\tau)$, where $\sigma$ is the adapter's scope-assignment function. The local retriever $R_{\mathrm{local}}$ collects up to $k_w$ relevant neighbors:
\begin{equation}
 \mathcal N_{u,s}=R_{\mathrm{local}}(\tau,\mathcal E_{u,s};k_w).
 \label{eq:local-neighborhood}
\end{equation}
The LLM-based update operator $U_{\mathrm{local}}$ uses this evidence and neighborhood to edit the collection:
\begin{equation}
 (\mathcal E'_{u,s},\Delta_{u,s})
 =U_{\mathrm{local}}(\mathcal E_{u,s},\tau;\mathcal N_{u,s}),
 \label{eq:local-update}
\end{equation}
Primes denote updated collections. The edit set $\Delta_{u,s}$ contains accepted additions, replacements, or removals; a no-op leaves it empty. The manager is instructed to abstract source-specific values into semantic slots and preference-retrieval guidance. Retrieved neighbors support consolidation with existing procedures; an empty neighborhood still permits additions.

For each edit $\delta\in\Delta_{u,s}$, $R_{\mathrm{shared}}$ retrieves up to $k_w$ shared neighbors. Their union supplies the shared update operator $U_{\mathrm{shared}}$:
\begin{equation}
 \mathcal N_s^{\mathrm{shared}}=\bigcup_{\delta\in\Delta_{u,s}}R_{\mathrm{shared}}(\delta,\mathcal R_s;k_w),
 \label{eq:shared-neighborhood}
\end{equation}
\begin{equation}
 \mathcal R'_s=U_{\mathrm{shared}}(\mathcal R_s,\Delta_{u,s};\mathcal N_s^{\mathrm{shared}}).
 \label{eq:shared-update}
\end{equation}
The shared manager follows the same abstraction rules but decides edits against its own collection. A local replacement or removal need not induce the same shared operation; without accepted local edits, shared memory is unchanged.

\subsection{Joint experience retrieval and active preference retrieval}
\label{sec:experience-read}

\paragraph{Scope-first experience retrieval.}
For task $x$ from user $u$, the adapter constructs $J$ scope groups $\mathbf G(x)=(G_0,\ldots,G_{J-1})$, from direct matches to broader fallbacks. Each $G_j$ contains scopes at priority $j$, with smaller $j$ indicating closer matches. Its local--shared candidate set is
\begin{equation}
 C_j=\bigcup_{s\in G_j}\left[R(x,\mathcal E_{u,s};K_c)\cup R(x,\mathcal R_s;K_c)\right],
\end{equation}
where the experience retriever $R$ returns up to $K_c$ candidates per collection. The guidance set $H_x$ is
\begin{equation}
 H_x=\operatorname{ScopeTopK}\!\left(x,(C_0,\ldots,C_{J-1}),K\right).
 \label{eq:scope-retrieval}
\end{equation}
Within each group, $\operatorname{ScopeTopK}$ jointly reranks both sources under a single $K$-entry budget. Lower-priority groups only fill remaining slots and cannot displace earlier selections. Selection stops at $K$ entries or candidate exhaustion. Appendix~\ref{app:retrieval-policy} specifies benchmark-specific candidate filtering and duplicate handling.

\paragraph{Active preference retrieval.}
For tasks with a preference channel, $R_{\mathrm{pref}}(q,\mathcal P_u;k_p)$ retrieves at most $k_p$ entries for query $q$. Initial recall $B_u^0=R_{\mathrm{pref}}(x,\mathcal P_u;k_p)$ joins $x$ and $H_x$ in message context $h_0$. At step $t$, LLM policy $\pi_\theta$ with parameters $\theta$ selects action $a_t$ from context $h_t$:
\begin{equation}
 a_t \sim \pi_\theta(\,\cdot\mid h_t), \qquad
 a_t \in \mathcal A_{\mathrm{env}}\cup\mathcal A_{\mathrm{pref}}\cup\mathcal A_{\mathrm{out}},
 \label{eq:agent-memory-action}
\end{equation}
The action spaces represent environment interaction, preference retrieval, and terminal responses, respectively. The policy decides whether and what to retrieve; experiences guide this decision without automatically triggering it. A preference action $a_t\in\mathcal A_{\mathrm{pref}}$ with query $q_t$ returns observation
\begin{equation}
 o_t=R_{\mathrm{pref}}(q_t,\mathcal P_u;k_p).
 \label{eq:preference-retrieval}
\end{equation}
The environment fixes $u$, preventing cross-user preference access. Read-only retrieval appends the action and its observation to the context:
\begin{equation}
 h_{t+1}=h_t\mathbin{\Vert}(a_t,o_t),
 \label{eq:preference-context-update}
\end{equation}
where $\Vert$ denotes context concatenation. The agent can retrieve again or act within the execution budget. Experiences guide which constraints to resolve, while current task requirements and personal preferences take precedence.

\section{Experiments}
\label{sec:experiments}

\subsection{Experimental Setup}
\label{sec:setup}

\paragraph{Benchmarks and metrics.}
\textbf{Mind2Web} evaluates instruction-following web agents on real-world websites, with human-demonstrated action sequences and generalization across tasks, websites, and domains \citep{deng2023mind2web}. We report task-macro element accuracy, action F1, and step success, measuring element selection, action prediction, and their joint correctness.
\textbf{VitaBench~2.0} evaluates personalized and proactive behavior in temporally ordered user interactions across delivery, in-store consumption, and travel services \citep{chen2026vitabench}. Tasks require tracking evolving preferences and acquiring missing information, with rubric-based evaluation of task execution. We report Avg@3, Pass@3, and Pass$^3$: mean success over three trials, success in at least one trial, and success in all three, respectively. All scores use the same 771 subtasks; success requires full reward, and missing or error outcomes count as failures.
\textbf{MemoryCode} is a multi-session coding benchmark that tests whether models retain and apply coding instructions amid irrelevant information and changing requirements \citep{rakotonirina2025tools}. Generated code is assessed against the applicable rules using the original AST/regex evaluator. We report dialogue-macro reward on a 0--100 scale, denoted D-MS.

\paragraph{Baselines.}
We construct five controlled memory configurations.
\textsc{None} disables long-term memory; \textsc{Pref} uses only the current user's personal preferences. \textsc{Local} adds user-local reusable experiences. \textsc{Share-all} additionally shares experiences and retrieves other users' preferences as source-labeled, non-authoritative hints, reserving slots for the current user. \textsc{ShareMem} shares reusable experiences while keeping preference retrieval user-bound; \textsc{Local} is its counterpart without cross-user sharing. Mind2Web omits \textsc{Pref} and \textsc{Share-all} because it has no preference channel. We additionally compare with benchmark-specific references: AWM \citep{wang2024agent} on Mind2Web; RAG, Rewrite, Full-CTX, and oracle Ground-truth on VitaBench~2.0; and Full-CTX and Full-Pref on MemoryCode.

\paragraph{Models.}
We use Qwen 3.6-27B \citep{qwen3.6-27b}, DeepSeek v4-flash \citep{deepseekai2026deepseekv4}, Gemma 4-26B \citep{gemmateam2026gemma4}, and Gemini 3 Flash \citep{googledeepmind2025gemini3flash} as foundation models. Retrieval uses BGE-M3 embeddings and BGE-Reranker-v2-M3 \citep{li2023making,chen2024bge}.

\paragraph{Implementation details.}
Scopes are supplied by website metadata in Mind2Web, task domains in VitaBench~2.0, and a single coding scope in MemoryCode. Mind2Web and MemoryCode construct experience collections offline from designated training trajectories or historical sessions and keep them fixed during evaluation. VitaBench~2.0 starts with empty experience collections and updates them online from completed full-reward subtasks, making accepted updates available to subsequent tasks. We set $k_w=5$, $K=5$, $k_p=10$, and $K_c=25$. Appendix~\ref{app:details} details benchmark protocols and score aggregation; Appendices~\ref{app:memory-configurations} and~\ref{app:retrieval-policy} specify memory-access configurations and update/retrieval procedures, respectively.

\subsection{Main Results}
\label{sec:results}
\begin{table}[h]
\centering
\caption{\textbf{Mind2Web.}
Three-seed task-macro scores (\%). EA: element accuracy; Act.\ F1: action F1; Step: step success.
$\Delta$: \textsc{ShareMem} $-$ \textsc{Local} (pp). Bold marks the best score.}
\label{tab:m2w-main}
{\small
\setlength{\tabcolsep}{2.7pt}
\begin{tabular*}{\textwidth}{@{\extracolsep{\fill}}l rrr rrr rrr rrr@{}}
\toprule
 & \multicolumn{3}{c}{Qwen 3.6-27B} & \multicolumn{3}{c}{DeepSeek v4-flash} & \multicolumn{3}{c}{Gemma 4-26B} & \multicolumn{3}{c}{Gemini 3 Flash} \\
\cmidrule(lr){2-4}\cmidrule(lr){5-7}\cmidrule(lr){8-10}\cmidrule(lr){11-13}
Memory & EA & Act.\ F1 & Step & EA & Act.\ F1 & Step & EA & Act.\ F1 & Step & EA & Act.\ F1 & Step \\
\midrule
\textcolor{scoregray}{None}   & \textcolor{scoregray}{42.40} & \textcolor{scoregray}{48.70} & \textcolor{scoregray}{36.90}
 & \textcolor{scoregray}{42.66} & \textcolor{scoregray}{50.17} & \textcolor{scoregray}{36.77}
 & \textcolor{scoregray}{39.23} & \textcolor{scoregray}{47.13} & \textcolor{scoregray}{32.22}
 & \textcolor{scoregray}{43.12} & \textcolor{scoregray}{49.04} & \textcolor{scoregray}{38.12} \\
\textsc{Local}  & 47.07 & 56.81 & 43.06
 & 48.12 & 58.35 & 44.17
 & 45.22 & 56.68 & 40.67
 & 50.14 & 57.13 & 45.29 \\
\rowcolor{orange!10}\textsc{ShareMem} & \textbf{51.32} & \textbf{58.63} & \textbf{47.16}
 & \textbf{51.80} & \textbf{59.35} & \textbf{47.96}
 & \textbf{47.05} & \textbf{58.20} & \textbf{43.13}
 & \textbf{51.87} & \textbf{58.35} & \textbf{47.64} \\
\rowcolor[gray]{0.94}\textcolor{scoregreen}{$\Delta$ S$-$L} & \textcolor{scoregreen}{\textbf{+4.25}} & \textcolor{scoregreen}{\textbf{+1.82}} & \textcolor{scoregreen}{\textbf{+4.10}}
 & \textcolor{scoregreen}{\textbf{+3.68}} & \textcolor{scoregreen}{\textbf{+1.00}} & \textcolor{scoregreen}{\textbf{+3.79}}
 & \textcolor{scoregreen}{\textbf{+1.83}} & \textcolor{scoregreen}{\textbf{+1.52}} & \textcolor{scoregreen}{\textbf{+2.46}}
 & \textcolor{scoregreen}{\textbf{+1.73}} & \textcolor{scoregreen}{\textbf{+1.22}} & \textcolor{scoregreen}{\textbf{+2.35}} \\
\bottomrule
\end{tabular*}
}
\end{table}

On Mind2Web, Table~\ref{tab:m2w-main} shows that \textsc{Local} outperforms \textsc{None} across all four backbones, demonstrating the value of accumulated experience for web interaction. \methodname further improves all three metrics, increasing step success by 2.35--4.10 percentage points over \textsc{Local} and by approximately 10.5 points over \textsc{None} on average across backbones. The accompanying gains in element accuracy and action F1 indicate improvements in both webpage-element selection and action prediction. Since this evaluation excludes same-site experience from the receiving user's local history, the gains over \textsc{Local} support the value of accessing relevant experience from other users. Section~\ref{sec:analysis} examines how this benefit changes as local experience coverage increases.

\begin{table}[h]
\centering
\caption{\textbf{VitaBench~2.0.} Subtask-level scores (\%) over three trials. Bold marks the best score; $^\dagger$ marks a deficit to \textsc{Share-all} below 0.5 pp. $\Delta$: \textsc{ShareMem} $-$ \textsc{Local} (pp).}
\label{tab:vita-main}
{\small
\setlength{\tabcolsep}{1.6pt}
\begin{tabular*}{\textwidth}{@{\extracolsep{\fill}}l rrr rrr rrr rrr@{}}
\toprule
 & \multicolumn{3}{c}{Qwen 3.6-27B} & \multicolumn{3}{c}{DeepSeek v4-flash} & \multicolumn{3}{c}{Gemma 4-26B} & \multicolumn{3}{c}{Gemini 3 Flash} \\
\cmidrule(lr){2-4}\cmidrule(lr){5-7}\cmidrule(lr){8-10}\cmidrule(lr){11-13}
Condition & Avg@3 & Pass@3 & Pass$^3$ & Avg@3 & Pass@3 & Pass$^3$ & Avg@3 & Pass@3 & Pass$^3$ & Avg@3 & Pass@3 & Pass$^3$ \\
\midrule
\textcolor{scoregray}{None} & \textcolor{scoregray}{6.96} & \textcolor{scoregray}{13.75} & \textcolor{scoregray}{1.69}
 & \textcolor{scoregray}{7.05} & \textcolor{scoregray}{12.97} & \textcolor{scoregray}{2.33}
 & \textcolor{scoregray}{5.53} & \textcolor{scoregray}{10.89} & \textcolor{scoregray}{1.82}
 & \textcolor{scoregray}{8.69} & \textcolor{scoregray}{15.43} & \textcolor{scoregray}{3.24} \\
\textsc{Pref} & 26.93 & 46.82 & 9.99
 & 22.78 & 41.50 & 7.65
 & 17.51 & 34.76 & 3.89
 & 20.88 & 42.41 & 4.41 \\
\textsc{Local} & 26.93 & 47.21 & 9.34
 & 22.57 & 39.95 & 7.65
 & 18.33 & 36.71 & 3.37
 & 21.88 & 44.10 & 5.06 \\
\textsc{Share-all} & 25.90 & 46.04 & 8.43
 & 23.99 & \textbf{42.80} & 7.52
 & 18.24 & 36.84 & \textbf{4.41}
 & 22.96 & 42.93 & \textbf{5.84} \\
\rowcolor{orange!10}\textsc{ShareMem} & \textbf{29.10} & \textbf{49.94} & \textbf{11.28}
 & \textbf{25.03} & 42.67$^{\dagger}$ & \textbf{9.08}
 & \textbf{19.67} & \textbf{39.82} & 4.02$^{\dagger}$
 & \textbf{23.13} & \textbf{45.01} & 4.93 \\
\rowcolor[gray]{0.94}\textcolor{scoregreen}{$\Delta$ S$-$L} & \textcolor{scoregreen}{\textbf{+2.16}} & \textcolor{scoregreen}{\textbf{+2.72}} & \textcolor{scoregreen}{\textbf{+1.95}}
 & \textcolor{scoregreen}{\textbf{+2.46}} & \textcolor{scoregreen}{\textbf{+2.72}} & \textcolor{scoregreen}{\textbf{+1.43}}
 & \textcolor{scoregreen}{\textbf{+1.34}} & \textcolor{scoregreen}{\textbf{+3.11}} & \textcolor{scoregreen}{\textbf{+0.65}}
 & \textcolor{scoregreen}{\textbf{+1.25}} & \textcolor{scoregreen}{\textbf{+0.91}} & \textcolor{scorered}{$-$0.13} \\
\bottomrule
\end{tabular*}
}
\end{table}
On VitaBench~2.0, Table~\ref{tab:vita-main} shows that access to personal preferences provides the largest gain over \textsc{None}, whereas adding local experiences yields only small, inconsistent changes. Sharing experience provides a more consistent additional benefit: \methodname achieves the highest Avg@3 across all four backbones, outperforming \textsc{Local} by approximately 1.8 percentage points on average. Extending access to other users' preferences does not improve this average performance, as \textsc{Share-all} trails \methodname on Avg@3 for every backbone. This advantage, however, does not hold uniformly across metrics: \textsc{Share-all} achieves the highest Pass@3 for DeepSeek and Pass$^3$ for Gemma and Gemini, while \methodname falls slightly below \textsc{Local} on Gemini's Pass$^3$. Thus, sharing improves average task success more consistently than repeatability.

\begin{table}[h]
\centering
\caption{\textbf{MemoryCode.} Mean D-MS (0--100) over three fixed 20-dialogue subsets. Bold marks the best mean; $\Delta$: \textsc{ShareMem} $-$ \textsc{Local} (pp). The benchmark protocols and aggregation subsection of Appendix~\ref{app:details} specifies subset construction and dialogue-macro scoring.}
\label{tab:mc-main}
{\small
\setlength{\tabcolsep}{3pt}
\begin{tabular}{@{}lrrrr>{\columncolor{orange!10}}rr@{}}
\toprule
Backbone & \textcolor{scoregray}{None} & \textsc{Pref} & \textsc{Local} & \textsc{Share-all} & \textsc{ShareMem} & \textcolor{scoregreen}{$\Delta$ S$-$L} \\
\midrule
Qwen 3.6-27B & \textcolor{scoregray}{2.58} & 59.84 & 66.41 & 67.38 & \textbf{69.32} & \textcolor{scoregreen}{\textbf{+2.91}} \\
DeepSeek v4-flash & \textcolor{scoregray}{1.39} & 77.46 & 78.37 & 77.21 & \textbf{78.53} & \textcolor{scoregreen}{\textbf{+0.16}} \\
Gemma 4-26B & \textcolor{scoregray}{2.14} & 56.91 & 56.24 & 50.05 & \textbf{57.63} & \textcolor{scoregreen}{\textbf{+1.39}} \\
Gemini 3 Flash & \textcolor{scoregray}{1.49} & 64.12 & 68.07 & 62.98 & \textbf{69.01} & \textcolor{scoregreen}{\textbf{+0.94}} \\
\bottomrule
\end{tabular}
}
\end{table}
On MemoryCode, Table~\ref{tab:mc-main} shows a large improvement from \textsc{None} to \textsc{Pref}, highlighting the importance of access to personal coding requirements. Sharing reusable experience provides an additional benefit: \methodname achieves higher mean D-MS than \textsc{Local} across all four backbones, although the gain for DeepSeek is marginal. Broader preference access is less effective: \textsc{Share-all} scores below \methodname for every backbone and below \textsc{Pref} for Gemma and Gemini. Together, these results support sharing reusable experience while keeping preference retrieval bound to the receiving user's own coding requirements.

\begin{figure}[h]
\centering
\includegraphics[width=\linewidth]{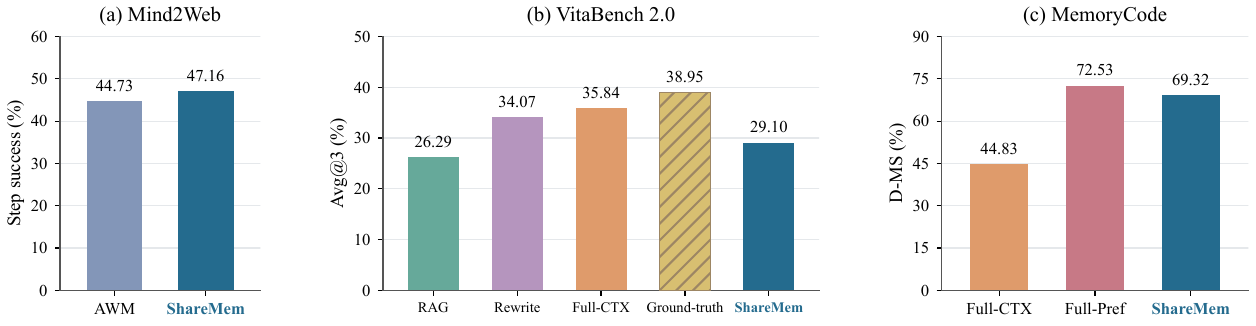}
\caption{\textbf{Benchmark-specific reference comparisons with Qwen.} Three-run means. Full-CTX supplies raw interaction history; Full-Pref supplies all extracted personal preferences without shared experiences or active retrieval.}
\label{fig:reference-comparison}
\end{figure}
Figure~\ref{fig:reference-comparison} complements the controlled comparison with \textsc{Local} by evaluating \methodname against benchmark-specific reference systems. On Mind2Web, \methodname outperforms AWM~\citep{wang2024agent} by 2.43 percentage points in step success. On VitaBench~2.0, it exceeds RAG with BGE-M3 retrieval but trails history rewriting and full-context conditioning, indicating that selective access does not capture all the benefits of richer historical context. Ground-truth supplies oracle preferences as an additional reference. On MemoryCode, \methodname substantially outperforms raw-history conditioning but remains 3.21 D-MS points below access to the complete extracted preference collection. These system-level comparisons highlight both the value and the limits of selective memory: concise experiences and preferences can be more effective than raw history, while broader access to relevant context or personal requirements can still provide additional gains.

\subsection{Ablation Studies}
\label{sec:write-analysis}

We examine the contributions of experience consolidation, scope-first retrieval, and active preference retrieval. For each comparison, we clarify changes to memory construction, retrieval policy, or preference access to contextualize the observed performance differences.

\paragraph{Two-stage consolidation.}
We first examine whether consolidating local experience before sharing improves the resulting shared memory. We compare \methodname with direct shared-memory writing, which processes raw sessions instead of accepted local edits. Both configurations retain an editable shared manager. Local pools are rebuilt across two-stage runs, so this comparison evaluates the writing procedures without holding local memory fixed. Two-stage consolidation produces a smaller shared experience pool (Figure~\ref{fig:write-pool-dynamics}) while achieving higher D-MS on all three subsets (Table~\ref{tab:write-main}). It also reduces shared-memory induction token usage (Table~\ref{tab:write-cost} in Appendix~\ref{app:write}). Together, these results support consolidating experience locally before sharing: the resulting shared memory combines a more compact pool and lower construction overhead with better downstream performance.

\begin{table}[h]
\centering
\begin{minipage}[t]{0.54\linewidth}
\vspace{0pt}
\caption{\textbf{Write-path ablations.}  Experiments conducted on MemoryCode with Qwen. D-MS (\%, dialogue-macro). 2S: two-stage; D: direct; A: append-only. Pools are rebuilt for each configuration.}
\label{tab:write-main}
\centering
{\small
\setlength{\tabcolsep}{2pt}
\begin{tabular}{@{}lrrr@{\hspace{6pt}}rrr@{}}
\toprule
 & \multicolumn{3}{c}{D-MS $\uparrow$} & \multicolumn{3}{c}{Global entries} \\
\cmidrule(lr){2-4}\cmidrule(lr){5-7}
Subset & 2S & D & A & 2S & D & A \\
\midrule
42 & \textbf{69.34} & 64.42 & 62.74 & 40 & 67 & 462 \\
23 & \textbf{66.94} & 66.75 & 65.70 & 17 & 79 & 467 \\
7 & \textbf{71.67} & 65.68 & 68.60 & 1 & 82 & 493 \\
\bottomrule
\end{tabular}
\par}
\end{minipage}\hfill
\begin{minipage}[t]{0.44\linewidth}
\vspace{0pt}
\centering
\includegraphics[width=\linewidth]{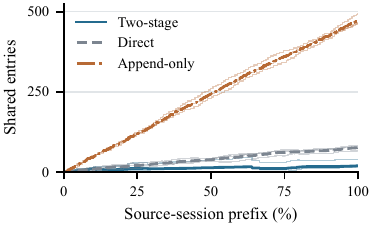}
\makeatletter\def\@captype{figure}\makeatother
\vspace{-7mm}
\caption{\textbf{Experience-pool dynamics.}}
\label{fig:write-pool-dynamics}
\end{minipage}
\vskip -0.1in
\end{table}

\paragraph{Editable consolidation.}
We next examine whether shared memory benefits from revising existing experience rather than continually accumulating new entries. \methodname supports adding, replacing, and removing entries, or leaving memory unchanged. The append-only variant adds newly extracted experiences with exact-string deduplication but neither revises nor removes existing entries. It also omits the existing pool from the extraction context, so this ablation changes both experience extraction and memory maintenance. Append-only storage yields lower D-MS on all three subsets (Table~\ref{tab:write-main}) despite producing substantially larger shared pools (Figure~\ref{fig:write-pool-dynamics}). Under the fixed five-entry retrieval budget, larger pools may increase competition among near-duplicate candidates without providing more useful guidance. Together, these results favor pool-aware, editable consolidation for maintaining compact and effective shared memory.

\begin{figure}[h]
\centering
\includegraphics[width=\linewidth]{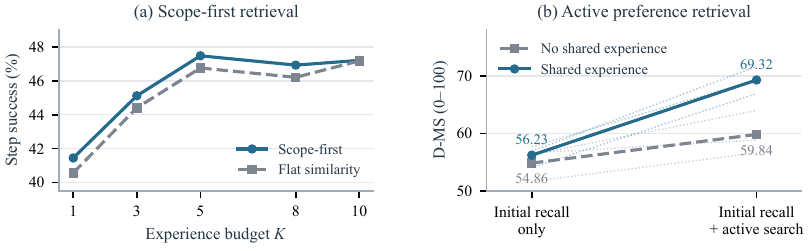}
\vspace{-7mm}
\caption{\textbf{Retrieval ablations.} (a) Mind2Web, scope-first retrieval outperforms flat retrieval across different experience budgets $K$. (b) Qwen, MemoryCode: three-subset means and dotted subset curves compare preference-only access with local and shared experience guidance, under initial recall alone or initial plus active recall.}
\label{fig:scope-retrieval}
\label{fig:private-mechanism}
\end{figure}
\paragraph{Scope-first retrieval.}
We examine whether scope information improves experience selection beyond similarity alone. Flat search jointly ranks all accessible local and shared experiences across scopes, without scope filtering or prioritization. We compare it with scope-first retrieval using frozen Mind2Web pools, the same similarity function, and matched retrieval budgets. In the Qwen sweep, scope-first retrieval achieves higher step success across the tested budgets (Figure~\ref{fig:scope-retrieval}(a)). The accompanying retrieval analysis shows a consistently larger fraction of entries from the task's website (Appendix~\ref{app:budget}). Although this fraction decreases for both strategies as the budget grows, scope-first retrieval retains substantially higher same-site coverage. Together, these results support scope priority as a way to improve contextual alignment with corresponding performance gains.

\paragraph{Active preference retrieval.}
\label{sec:private-analysis}
We test whether experience guidance and active preference retrieval provide complementary benefits on MemoryCode. With experience pools frozen, we compare initial recall alone with initial recall plus agent-initiated retrieval under two conditions: \textsc{Pref}, which provides only personal preferences, and \methodname, which additionally provides local and shared experience guidance. Relative to \textsc{Pref}, \methodname improves mean D-MS by 9.48 percentage points with active retrieval, compared with 1.37 points under initial recall alone (Figure~\ref{fig:private-mechanism}(b)). This larger gain holds across all three subsets, supporting complementarity between reusable guidance and access to user-specific constraints during execution. The comparison measures the combined contribution of local and shared experience, rather than the incremental benefit of sharing alone. Moreover, disabling active retrieval restricts both information access and interaction opportunities, so these results do not isolate retrieval frequency as the source of improvement.

\subsection{Further Analysis}
\label{sec:analysis}
\begin{figure}[h]
\centering
\begin{minipage}[t]{0.48\linewidth}
\centering
\includegraphics[width=\linewidth]{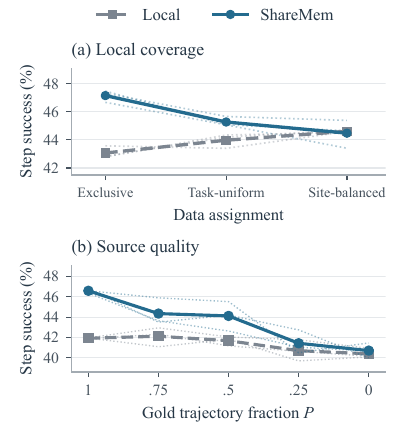}
\vspace{-7mm}
\caption{\textbf{Experience coverage and quality.} (a) Step success under different user assignments; thick curves show three-seed means and faint dotted curves individual seeds. (b) Replacing gold demonstrations with cached agent rollouts at fixed training-task coverage.}
\label{fig:reachability}
\label{fig:mixed-source}
\end{minipage}\hfill
\begin{minipage}[t]{0.48\linewidth}
\centering
\includegraphics[width=\linewidth]{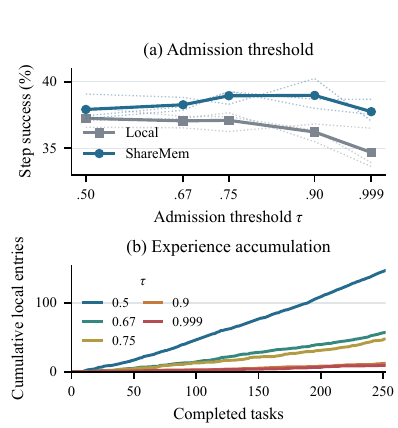}
\vspace{-7mm}
\caption{\textbf{Admission and accumulation.} Qwen, cold-start online Mind2Web; three-seed means. (a) Task-macro step success versus admission threshold $\tau$; dotted curves show individual seeds. (b) Total local experience entries across users versus completed tasks.}
\label{fig:admission-dynamics}
\end{minipage}
\end{figure}

\paragraph{Local experience coverage.}
We examine how the benefit of sharing changes with the availability of relevant local experience. Keeping the Mind2Web training and evaluation tasks fixed, we vary their assignment to users. Exclusive assigns each website's training trajectories to one user and its evaluation tasks to another. Task-uniform assigns tasks through a seeded hash, while Site-balanced distributes them round-robin within each website using a shared user offset across splits. Across three seeds, these assignments produce mean same-site local coverage of 0.00\%, 49.60\%, and 98.81\%, respectively (Table~\ref{tab:assignment-coverage}). As coverage increases, the corresponding mean gains of \methodname over \textsc{Local} in task-macro step success decline from 4.10 to 1.30 and $-0.13$ percentage points (Figure~\ref{fig:reachability}(a)). These results show that sharing is most useful in this setting when it supplies relevant experience missing from the recipient's history; its incremental benefit largely disappears when that experience is already available locally.

\paragraph{Experience quality.}
We next examine sensitivity to evidence quality by replacing gold demonstrations with cached agent rollouts while holding training-task coverage fixed. This intervention changes the evidence used for both experience construction and attached demonstrations. Moving from all-gold to all-rollout evidence reduces mean task-macro step success by 5.90 percentage points for \methodname, compared with 1.53 points for \textsc{Local}, across three seeds (Figure~\ref{fig:mixed-source}(b)). The decline in \methodname holds in every seed. Under this split, \textsc{Local} retrieves only cross-site experience, whereas shared guidance is usually same-site; the smaller \textsc{Local} decline therefore does not establish greater robustness to unreliable evidence. The asymmetry is consistent with source quality becoming more consequential when sharing supplies experience relevant to the current task. These results highlight the need to consider evidence reliability alongside experience coverage when constructing shared memory.

\paragraph{Admission thresholds.}
We next examine how admission filtering balances rollout quality against experience coverage. In cold-start online Mind2Web, agents begin with empty experience pools and update them from rollouts on evaluation tasks; admitted experience becomes available to subsequent tasks. A rollout is admitted only when its step-success rate reaches $\tau$, with $\tau \ge 0.5$ in this comparison. Stricter thresholds slow local experience accumulation (Figure~\ref{fig:admission-dynamics}). As $\tau$ increases from $0.5$ to $0.999$, \textsc{Local} loses approximately 2.6 percentage points in step success, while \methodname changes less overall but declines from its intermediate-threshold peak. Online scores also remain substantially below the offline results, consistent with reliance on noisier agent-generated evidence and limited experience during cold start. Unlike the source-quality sweep, which preserves training-task coverage, admission filtering can exclude evidence altogether. Together, these results highlight a trade-off: stricter filtering raises the success requirement for admitted rollouts but can leave agents without useful guidance. Effective shared memory therefore requires attention to both evidence quality and experience coverage.

\begin{figure}[!h]
\centering
\begin{minipage}[t]{0.49\linewidth}
\vspace{0pt}
\centering
\includegraphics[width=\linewidth]{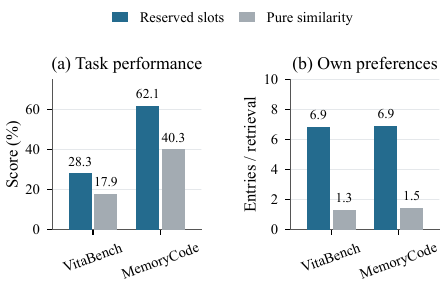}
\caption{\textbf{Preference-slot reservation.} We compare two retrieval settings: reserved preference slots versus pure similarity ranking. \textbf{(a)} Strict success on VitaBench~2.0 and D-MS on MemoryCode under the two settings. \textbf{(b)} Mean number of current-user entries per top-10 retrieval.}
\label{fig:preference-reservation}
\end{minipage}\hfill
\begin{minipage}[t]{0.49\linewidth}
\vspace{0pt}
\centering
\includegraphics[width=\linewidth]{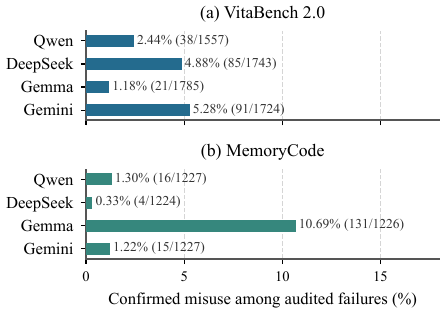}
\vspace{-7mm}
\caption{\textbf{Cross-user preference interference.} \textsc{Share-all}, pooled over three seeds. Labels show confirmed misuse rates and counts among failures exposed to other users' preferences. Confirmation requires unanimous agreement across three repeated model judgments.}
\label{fig:preference-interference}
\end{minipage}
\vskip -0.1in
\end{figure}
\paragraph{Cross-user preference interference.}
\textsc{Share-all} reserves seven of ten preference slots for the current user. Removing this reservation and ranking all preferences by similarity reduces current-user entries to fewer than two on average and lowers performance (Figure~\ref{fig:preference-reservation}). Yet reservation does not eliminate misuse: an audit of failed outputs identifies agents applying other users' values despite source labels, explicit user identification, and instructions against such use (Figure~\ref{fig:preference-interference}). These findings distinguish preserving preference access from ensuring correct application, motivating user-bound retrieval that restricts the preference channel to the current user's own store.

\section{Conclusion}
\label{sec:conclusion}

\methodname enables cross-user experience reuse through two-stage consolidation and preference retrieval bound to the receiving user. Our experiments show that sharing adds value beyond user-local memory, particularly when relevant experience is unavailable locally. Effective transfer nevertheless depends on experience quality, consolidation, and access to personal constraints during execution. Larger memory pools and broader access to other users' preferences do not necessarily improve performance, underscoring the distinction between making information accessible and making it applicable. Together, these findings support a practical principle for collective learning among LLM agents: share reusable guidance while grounding its application in the receiving user's own requirements.
\subsection*{AI use statement}  
Generative AI tools, specifically GPT-5, GPT-6, and Codex, were used for language polishing, structural reorganization, and retrieval/discovery of related work. They were not used to conduct experiments, generate numerical results, or produce scientific findings. All AI-assisted outputs, including suggested references, were independently verified by the authors. We take full responsibility for the final content of this work, including all text, claims, and artifacts.

\subsection*{Ethics statement}

Our experiments use public benchmarks. \methodname separates reusable experiences from user-specific preferences. Real-world deployment should require appropriate consent and safeguards against unintended cross-user information sharing.

\subsection*{Reproducibility statement}
Section~\ref{sec:method} specifies the memory organization, two-stage consolidation, and experience and preference retrieval procedures. Section~\ref{sec:setup} describes the benchmarks, evaluation metrics, backbone models, and retrieval settings. Appendix~\ref{app:details} documents benchmark adaptations, user and task assignments, seeded subset selection, offline and online evaluation protocols, and score aggregation, together with the memory-access configurations and update procedures. Appendix~\ref{app:additional-evidence} provides detailed ablation protocols, supplementary results, memory-construction costs, and the procedures used to audit cross-user preference interference. Appendix~\ref{app:prompts} includes key prompts for experience extraction, local and shared memory updates, and preference-aware execution.

\bibliography{iclr2027_conference}

@article{shinn2023reflexion,
  title={Reflexion: Language agents with verbal reinforcement learning},
  author={Shinn, Noah and Cassano, Federico and Gopinath, Ashwin and Narasimhan, Karthik and Yao, Shunyu},
  journal={Advances in neural information processing systems},
  volume={36},
  pages={8634--8652},
  year={2023}
}

@inproceedings{zhao2024expel,
  title={Expel: Llm agents are experiential learners},
  author={Zhao, Andrew and Huang, Daniel and Xu, Quentin and Lin, Matthieu and Liu, Yong-Jin and Huang, Gao},
  booktitle={Proceedings of the AAAI Conference on Artificial Intelligence},
  volume={38},
  number={17},
  pages={19632--19642},
  year={2024}
}

@article{wang2024agent,
  title={Agent workflow memory},
  author={Wang, Zora Zhiruo and Mao, Jiayuan and Fried, Daniel and Neubig, Graham},
  journal={arXiv preprint arXiv:2409.07429},
  year={2024}
}

@article{rezazadeh2025collaborative,
  title={Collaborative memory: Multi-user memory sharing in llm agents with dynamic access control},
  author={Rezazadeh, Alireza and Li, Zichao and Lou, Ange and Zhao, Yuying and Wei, Wei and Bao, Yujia},
  journal={arXiv preprint arXiv:2505.18279},
  year={2025}
}

@article{wang2023voyager,
  title={Voyager: An open-ended embodied agent with large language models},
  author={Wang, Guanzhi and Xie, Yuqi and Jiang, Yunfan and Mandlekar, Ajay and Xiao, Chaowei and Zhu, Yuke and Fan, Linxi and Anandkumar, Anima},
  journal={arXiv preprint arXiv:2305.16291},
  year={2023}
}

@article{packer2023memgpt,
  title={Memgpt: Towards llms as operating systems},
  author={Packer, Charles and Wooders, Sarah and Lin, Kevin and Fang, Vivian and Patil, Shishir G and Stoica, Ion and Gonzalez, Joseph E},
  journal={arXiv preprint arXiv:2310.08560},
  year={2023}
}

@article{xu2026mem,
  title={A-mem: Agentic memory for llm agents},
  author={Xu, Wujiang and Liang, Zujie and Mei, Kai and Gao, Hang and Tan, Juntao and Zhang, Yongfeng},
  journal={Advances in Neural Information Processing Systems},
  volume={38},
  pages={17577--17604},
  year={2026}
}

@article{chhikara2025mem0,
  title={Mem0: Building production-ready ai agents with scalable long-term memory},
  author={Chhikara, Prateek and Khant, Dev and Aryan, Saket and Singh, Taranjeet and Yadav, Deshraj},
  journal={arXiv preprint arXiv:2504.19413},
  year={2025}
}

@article{deng2023mind2web,
  title={Mind2web: Towards a generalist agent for the web},
  author={Deng, Xiang and Gu, Yu and Zheng, Boyuan and Chen, Shijie and Stevens, Sam and Wang, Boshi and Sun, Huan and Su, Yu},
  journal={Advances in Neural Information Processing Systems},
  volume={36},
  pages={28091--28114},
  year={2023}
}

@article{chen2026vitabench,
  title={VitaBench 2.0: Evaluating Personalized and Proactive Agents in Long-Term User Interactions},
  author={Chen, Yuxin and Zhang, Yi and Cai, Zhengzhou and Shi, Yaorui and Yao, Zhiyuan and Cui, Chenhang and Zheng, Jingnan and Huo, Yaqi and Su, Xi and Gu, Qi and others},
  journal={arXiv preprint arXiv:2605.27141},
  year={2026}
}

@inproceedings{rakotonirina2025tools,
  title={From tools to teammates: evaluating LLMs in multi-session coding interactions},
  author={Rakotonirina, Nathana{\"e}l Carraz and Hamdy, Mohammed and Campos, Jon Ander and Weber, Lucas and Testoni, Alberto and Fadaee, Marzieh and Pezzelle, Sandro and Del Tredici, Marco},
  booktitle={Proceedings of the 63rd Annual Meeting of the Association for Computational Linguistics (Volume 1: Long Papers)},
  pages={19609--19642},
  year={2025}
}

@inproceedings{park2023generative,
  title={Generative agents: Interactive simulacra of human behavior},
  author={Park, Joon Sung and O'Brien, Joseph and Cai, Carrie Jun and Morris, Meredith Ringel and Liang, Percy and Bernstein, Michael S},
  booktitle={Proceedings of the 36th annual acm symposium on user interface software and technology},
  pages={1--22},
  year={2023}
}

@inproceedings{zhong2024memorybank,
  title={Memorybank: Enhancing large language models with long-term memory},
  author={Zhong, Wanjun and Guo, Lianghong and Gao, Qiqi and Ye, He and Wang, Yanlin},
  booktitle={Proceedings of the AAAI conference on artificial intelligence},
  volume={38},
  number={17},
  pages={19724--19731},
  year={2024}
}

@article{sumers2023cognitive,
  title={Cognitive architectures for language agents},
  author={Sumers, Theodore R and Yao, Shunyu and Narasimhan, Karthik and Griffiths, Thomas L},
  journal={arXiv preprint arXiv:2309.02427},
  year={2023}
}

@inproceedings{zhang2026agentic,
  title={Agentic context engineering: Evolving contexts for self-improving language models},
  author={Zhang, Qizheng and Hu, Changran and Upasani, Shubhangi and Ma, Boyuan and Hong, Fenglu and Kamanuru, Vamsidhar and Rainton, Jay and Wu, Chen and Ji, Mengmeng and Li, Hanchen and others},
  booktitle={International Conference on Learning Representations},
  volume={2026},
  pages={86069--86100},
  year={2026}
}

@inproceedings{ouyang2026reasoningbank,
  title={Reasoningbank: Scaling agent self-evolving with reasoning memory},
  author={Ouyang, Siru and Yan, Jun and Hsu, I and Chen, Yanfei and Jiang, Ke and Wang, Zifeng and Han, Rujun and Le, Long and Daruki, Samira and Tang, Xiangru and others},
  booktitle={International Conference on Learning Representations},
  volume={2026},
  pages={94327--94354},
  year={2026}
}

@article{yang2026federatedskill,
  title={FederatedSkill: Federated Learning for Agentic Skill Evolution},
  author={Yang, Jingbo and Yao, Guanyu and Zhang, Yang and Kompella, Ramana Rao and Liu, Gaowen and Chang, Shiyu},
  journal={arXiv preprint arXiv:2606.03143},
  year={2026}
}

@article{ma2026skillclaw,
  title={Skillclaw: Let skills evolve collectively with agentic evolver},
  author={Ma, Ziyu and Yang, Shidong and Ji, Yuxiang and Wang, Xucong and Wang, Yong and Hu, Yiming and Huang, Tongwen and Chu, Xiangxiang},
  journal={arXiv preprint arXiv:2604.08377},
  year={2026}
}

@article{wu2026scaling,
  title={Scaling Teams or Scaling Time? Memory Enabled Lifelong Learning in LLM Multi-Agent Systems},
  author={Wu, Shanglin and Luo, Yuyang and Liang, Yueqing and Shi, Kaiwen and Ye, Yanfang and Payani, Ali and Shu, Kai},
  journal={arXiv preprint arXiv:2604.03295},
  year={2026}
}

@article{hou2026fedworld,
  title={FedWorld: Scope-Aware Federation of Agent World Models},
  author={Hou, Yuchao},
  journal={arXiv preprint arXiv:2608.01561},
  year={2026}
}

@article{tang2025agent,
  title={Agent kb: Leveraging cross-domain experience for agentic problem solving},
  author={Tang, Xiangru and Qin, Tianrui and Peng, Tianhao and Zhou, Ziyang and Shao, Daniel and Du, Tingting and Wei, Xinming and Xia, Peng and Wu, Fang and Zhu, He and others},
  journal={arXiv preprint arXiv:2507.06229},
  year={2025}
}

@article{gao2024memory,
  title={Memory sharing for large language model based agents},
  author={Gao, Hang and Zhang, Yongfeng},
  journal={arXiv e-prints},
  pages={arXiv--2404},
  year={2024}
}

@inproceedings{liang2026learning,
  title={Learning how to remember: A meta-cognitive management method for structured and transferable agent memory},
  author={Liang, Sirui and Cao, Pengfei and Zhao, Jian and Teng, Wenhao and Liao, Xiangwen and Zhao, Jun and Liu, Kang},
  booktitle={Findings of the Association for Computational Linguistics: ACL 2026},
  pages={30733--30753},
  year={2026}
}

@inproceedings{salemi2024lamp,
  title={Lamp: When large language models meet personalization},
  author={Salemi, Alireza and Mysore, Sheshera and Bendersky, Michael and Zamani, Hamed},
  booktitle={Proceedings of the 62nd Annual Meeting of the Association for Computational Linguistics (Volume 1: Long Papers)},
  pages={7370--7392},
  year={2024}
}

@inproceedings{wang2024crafting,
  title={Crafting personalized agents through retrieval-augmented generation on editable memory graphs},
  author={Wang, Zheng and Li, Zhongyang and Jiang, Zeren and Tu, Dandan and Shi, Wei},
  booktitle={Proceedings of the 2024 conference on empirical methods in natural language processing},
  pages={4891--4906},
  year={2024}
}

@inproceedings{tan2025prospect,
  title={In prospect and retrospect: Reflective memory management for long-term personalized dialogue agents},
  author={Tan, Zhen and Yan, Jun and Hsu, I-Hung and Han, Rujun and Wang, Zifeng and Le, Long and Song, Yiwen and Chen, Yanfei and Palangi, Hamid and Lee, George and others},
  booktitle={Proceedings of the 63rd Annual Meeting of the Association for Computational Linguistics (Volume 1: Long Papers)},
  pages={8416--8439},
  year={2025}
}

@misc{gemmateam2026gemma4,
      title={Gemma 4 Technical Report}, 
      author={{Gemma Team}},
      year={2026},
      eprint={2607.02770},
      archivePrefix={arXiv},
      primaryClass={cs.CL},
      url={https://arxiv.org/abs/2607.02770}, 
}

@misc{qwen3.6-27b,
    title  = {{Qwen3.6-27B}: Flagship-Level Coding in a {27B} Dense Model},
    author = {{Qwen Team}},
    month  = {April},
    year   = {2026},
    url    = {https://qwen.ai/blog?id=qwen3.6-27b}
}

@misc{deepseekai2026deepseekv4,
      title={DeepSeek-V4: Towards Highly Efficient Million-Token Context Intelligence},
      author={DeepSeek-AI},
      year={2026},
}

@misc{li2023making,
      title={Making Large Language Models A Better Foundation For Dense Retrieval}, 
      author={Chaofan Li and Zheng Liu and Shitao Xiao and Yingxia Shao},
      year={2023},
      eprint={2312.15503},
      archivePrefix={arXiv},
      primaryClass={cs.CL}
}

@misc{chen2024bge,
      title={BGE M3-Embedding: Multi-Lingual, Multi-Functionality, Multi-Granularity Text Embeddings Through Self-Knowledge Distillation}, 
      author={Jianlv Chen and Shitao Xiao and Peitian Zhang and Kun Luo and Defu Lian and Zheng Liu},
      year={2024},
      eprint={2402.03216},
      archivePrefix={arXiv},
      primaryClass={cs.CL}
}

@misc{googledeepmind2025gemini3flash,
  author = {{Google DeepMind}},
  title = {Gemini 3 Flash: Frontier Intelligence Built for Speed},
  year = {2025},
  url = {https://blog.google/products-and-platforms/products/gemini/gemini-3-flash/}
}
\bibliographystyle{iclr2027_conference}

\appendix
\section{Implementation and Evaluation Protocols}
\label{app:details}

This appendix details the benchmark adaptations and evaluation protocols, the memory configurations used for comparison, and the procedures for memory construction and retrieval.

\subsection{Benchmark protocols and aggregation}
\paragraph{Offline and online protocols.}
We distinguish the protocols by when reusable experience collections are updated. In the offline protocol, experiences are constructed from designated demonstrations or historical sessions before evaluation and remain fixed while evaluation tasks are executed. In the online protocol, experience construction and evaluation are interleaved: completed interactions can update memory for subsequent tasks. Mind2Web and MemoryCode use the offline protocol in the main experiments; VitaBench~2.0 uses the online protocol.

\paragraph{Mind2Web.}
Mind2Web evaluates instruction-following web agents using natural-language tasks and human-demonstrated action sequences on real-world websites \citep{deng2023mind2web}.
We adapt web-navigation tasks to a population of 20 simulated users by assigning ownership to training demonstrations and evaluation tasks. Under the main Exclusive assignment, each website's training demonstrations belong to one user, while its evaluation tasks belong to a different user in the same population. A user may therefore possess local experience from other websites while lacking experience from the evaluation website. Local and shared experience collections are constructed from the assigned training demonstrations and frozen before evaluating 252 tasks. At each recorded step, the agent predicts a target element and action from the task instruction and webpage observation. This is evaluation on recorded webpage states rather than a live-browser rollout. Mind2Web provides no personal-preference channel in this adaptation. The data-assignment study changes task ownership, whereas the separate online admission study constructs experiences from agent rollouts starting with empty collections.

\paragraph{VitaBench~2.0.}
VitaBench~2.0 evaluates personalized and proactive assistance through temporally ordered user interactions and service tasks that require applying user preferences and acquiring missing information \citep{chen2026vitabench}.
We retain the benchmark's 56 users and 771 subtasks across delivery, in-store consumption, and travel services. Subtasks follow each user's temporal order, with execution interleaved across users. User interactions supplied before a subtask update that user's preference store; initial recall and agent-issued queries retrieve from the preferences available at that point. Local and shared experience collections start empty. After a subtask finishes, its trajectory becomes eligible for experience updates only if it receives full reward. Accepted shared updates become available to subsequent tasks once committed. Thus, preference acquisition from user interactions and experience acquisition from successful execution are separate processes. Each trial starts a new memory trajectory; memory is not carried between trials. The benchmark's rubric-based evaluator determines subtask success.

\paragraph{MemoryCode.}
MemoryCode evaluates whether agents retain and apply user-specific coding requirements across multiple sessions containing irrelevant information and changing instructions \citep{rakotonirina2025tools}.
We treat each dialogue as one user's multi-session history. Historical sessions supply local reusable experiences and, separately, a user-specific store of coding requirements. Accepted local experience updates provide evidence for shared consolidation. Evaluation uses the history queries associated with the latest evaluated session of each selected dialogue and the corresponding preference snapshot. Experience collections and preference stores are fixed during evaluation; agents may retrieve additional preferences while generating code, but evaluation outputs do not update these stores. Generated code is scored against the applicable requirements using the benchmark's AST/regex-based evaluator. Seeds 42, 23, and 7 select three fixed 20-dialogue subsets containing 395, 413, and 420 coding tasks, respectively. Conditions within a subset use the same dialogues and evaluation queries.

\paragraph{Metrics and aggregation.}
For Mind2Web, element accuracy, action F1, and step success are averaged within each task and then equally across tasks. Main scores retain skipped-step records with zero scores; task-paired analyses that exclude unevaluable steps specify their resulting sample size separately. The reported main scores average three runs on the same evaluation tasks. For VitaBench~2.0, success means full subtask reward. We align each subtask across three trials: Avg@3 averages its success indicators, Pass@3 records success in at least one trial, and Pass$^3$ requires success in all three. Each quantity is then averaged across subtasks. For MemoryCode, the benchmark evaluator aggregates applicable rule scores into a score for each generated output. Scored outputs are averaged within each dialogue, followed by an equal-weight average across dialogues and conversion to a 0--100 scale to obtain D-MS. Main results equally average the three subset-level D-MS scores. 

\subsection{Memory configurations}
\label{app:memory-configurations}
We construct controlled configurations by varying access to personal preferences, user-local experiences, and shared experiences. Table~\ref{tab:memory-configurations} summarizes the accessible information sources. Access denotes permission to retrieve from a collection, not a guarantee that it contains relevant entries or that the agent uses them.

\begin{table}[htbp]
\centering
\caption{\textbf{Accessible memory sources.} Preference columns apply to VitaBench~2.0 and MemoryCode. Local experiences belong to the current user; shared experiences are available across users under the applicable scopes.}
\label{tab:memory-configurations}
{\small
\begin{tabular}{lcccc}
\toprule
 & \multicolumn{2}{c}{Personal preferences} & \multicolumn{2}{c}{Reusable experiences} \\
\cmidrule(lr){2-3}\cmidrule(lr){4-5}
Configuration & Current user & Other users & Local & Shared \\
\midrule
\textsc{None} & -- & -- & -- & -- \\
\textsc{Pref} & \cmark & -- & -- & -- \\
\textsc{Local} & \cmark & -- & \cmark & -- \\
\textsc{Share-all} & \cmark & \cmark & \cmark & \cmark \\
\rowcolor{orange!10}\textsc{ShareMem} & \cmark & -- & \cmark & \cmark \\
\bottomrule
\end{tabular}
\par}
\end{table}

\paragraph{No memory and preference-only access.}
\textsc{None} disables long-term memory construction and retrieval while retaining the task input and the within-task context provided by the evaluation protocol. It does not remove information supplied directly in the current request. \textsc{Pref} enables only the current user's preference store, including initial task-query recall and agent-initiated retrieval during execution; it provides no reusable experiences.

\paragraph{Local and shared experience access.}
\textsc{Local} adds experiences extracted from the current user's interactions. It uses the same local experience-management procedure as \textsc{ShareMem}, but does not expose shared experiences. \textsc{ShareMem} additionally makes the shared experience collections accessible, while keeping preference retrieval bound to the current user. Local and shared experience candidates compete under a joint final budget of $K=5$.

\paragraph{Cross-user preference access.}
\textsc{Share-all} retains local and shared experience access and additionally permits retrieval of other users' preferences. Both initial recall and subsequent preference queries return at most $k_p=10$ entries. When sufficient distinct candidates exist, at least seven slots are reserved for the current user; other-user entries fill the remaining slots, with current-user entries filling unused capacity when available. The composition can differ when either source lacks candidates. Current-user preferences are presented as authoritative, whereas other-user preferences are source-labeled, non-authoritative hints. This condition tests preference sharing in addition to experience sharing, not exposure to unlabeled foreign information. 

\paragraph{Benchmark applicability and comparison controls.}
Mind2Web has no preference channel, so the controlled configurations are \textsc{None}, \textsc{Local}, and \textsc{ShareMem}. VitaBench~2.0 and MemoryCode use all five configurations. Comparisons within a benchmark and seed use the same backbone, evaluation tasks, and scoring procedure, with matched retrieval limits for enabled channels.

\subsection{Experience updates and retrieval}
\label{app:retrieval-policy}

\paragraph{Scope instantiation.}
Scopes describe the intended applicability of reusable experiences, independently of user ownership. The benchmark adapters instantiate the scope groups defined in the main method using supplied environment labels. Mind2Web stores experiences by website; website, subdomain, and domain metadata determine retrieval priority. Subdomains and domains group related website collections rather than introducing additional copies of each experience. VitaBench~2.0 organizes experiences by task domain, while MemoryCode uses a single coding scope. Table~\ref{tab:scope-adapters} summarizes these choices. Scope matching guides candidate selection; it does not guarantee that every experience within a matched collection applies to the current task.

\begin{table}[htbp]
\centering
\caption{\textbf{Scope instantiation and retrieval priorities.} Priorities run from left to right. Each scope can contain current-user local experiences and shared experiences.}
\label{tab:scope-adapters}
{\small
\begin{tabular}{@{}lll@{}}
\toprule
Benchmark & Collection scope & Matching priority \\
\midrule
Mind2Web & Website & Website $\rightarrow$ subdomain $\rightarrow$ domain $\rightarrow$ other sites \\
VitaBench~2.0 & Task domain & Same domain $\rightarrow$ other domains \\
MemoryCode & Coding & Single group; no broader-scope fallback \\
\bottomrule
\end{tabular}
\par}
\end{table}

\paragraph{Local experience updates.}
The local manager combines new interaction evidence with up to $k_w=5$ nearby experiences from $\mathcal E_{u,s}$ to update the collection. The evidence consists of designated demonstrations in offline Mind2Web, accepted completed subtasks in VitaBench~2.0, and historical sessions in MemoryCode. The manager can add an experience, replace existing content, remove an entry, or leave the collection unchanged. Its decision concerns the collection rather than a single predetermined entry. Experience formation and consolidation are performed together: new evidence may extend an existing procedure or support a new one, depending on the retrieved neighborhood. Personal values are intended to be replaced with guidance for obtaining the receiving user's applicable preferences.

\paragraph{Shared experience updates.}
Accepted local changes form $\Delta_{u,s}$, the evidence passed to the shared manager. This manager retrieves nearby entries from $\mathcal R_s$ and independently decides how to update the shared collection. A local replacement or removal is therefore not replayed as the same operation on shared memory: the two collections have different contents and consolidation contexts. 

\paragraph{Experience selection.}
Retrieval first collects up to $K_c=25$ candidates from each accessible local or shared collection. Within a scope-priority group, candidates compete by relevance without reserved local/shared quotas. The selector fills a joint budget of $K=5$ entries and consults lower-priority groups only when capacity remains; fallback candidates do not displace already selected higher-priority entries. It can return fewer entries when candidates are exhausted. Mind2Web applies the website-based hierarchy in Table~\ref{tab:scope-adapters}. VitaBench~2.0 first bounds local candidates through domain-prioritized retrieval, then combines them with shared candidates for final scope-first selection. Both scoped selectors suppress repeated text. MemoryCode instead jointly reranks local and shared candidates within its single coding scope, without a fallback hierarchy or an additional text-deduplication stage.

\paragraph{Preference construction.}
Personal preferences are stored as user-indexed textual entries, separately from reusable experiences. In VitaBench~2.0, a Mem0-based manager extracts and updates preference entries from the user interactions supplied before each subtask, retaining concrete user-specific values for subsequent retrieval. In MemoryCode, historical sessions are processed in order to maintain a collection of coding requirements for each user. The collection at the evaluation endpoint is retained as a fixed preference snapshot throughout evaluation. Mind2Web does not instantiate a preference store.

\paragraph{Preference retrieval.}
Where a preference channel is available, the task query initially retrieves up to $k_p=10$ entries. During execution, the agent can issue further queries to recover more specific requirements or resolve missing information. Each query returns at most $k_p$ entries and leaves the preference store unchanged. The agent chooses whether and what to retrieve from the task, selected experiences, and accumulated observations; retrieving an experience does not automatically trigger a preference query. MemoryCode permits this access to its fixed rule snapshot, while VitaBench~2.0 exposes preferences accumulated up to the current interaction. Access is bound to the current user except in the explicitly permissive \textsc{Share-all} configuration described above. Memory updates between interactions in the online protocol are distinct from these read-only retrieval actions.

\section{Supplementary Ablations and Mechanism Evidence}
\label{app:additional-evidence}

\subsection{Two-stage and editable consolidation}
\label{app:write}
We supplement the main write-path ablation with paired performance comparisons and shared-pool redundancy statistics. The Qwen MemoryCode comparisons match evaluation tasks, personal preference snapshots, and retrieval budgets. Direct writing changes the shared update path; append-only changes both local and shared maintenance. Experience pools are rebuilt, and append-only extraction omits existing-pool context.

\begin{table}[!htb]
\centering
\caption{\textbf{Task-paired effects of write-path changes.} Qwen, MemoryCode. Tasks are paired within each subset, then pooled over three subsets. Differences are alternative minus reference in task-weighted percentage points. Wilcoxon signed-rank tests are computed over the pooled task pairs.}
\label{tab:write-ablation}
\label{tab:append-ablation}
{\small
\begin{tabular}{llrr}
\toprule
Change & Configuration & Score difference & Wilcoxon $p$ \\
\midrule
Direct $-$ two-stage & \textsc{Share-all} & $-2.74$ & $.0011$ \\
\rowcolor{orange!10} & \textsc{ShareMem} & $-2.58$ & $.013$ \\
\midrule
Append-only $-$ editable & \textsc{Local} & $-2.00$ & $.002$ \\
 & \textsc{Share-all} & $-4.53$ & $1.2\times10^{-11}$ \\
\rowcolor{orange!10} & \textsc{ShareMem} & $-3.70$ & $1.5\times10^{-9}$ \\
\bottomrule
\end{tabular}
}
\end{table}

Table~\ref{tab:write-ablation} shows lower task-weighted scores under both alternative write paths. Direct writing affects configurations that access shared experiences; append-only additionally affects \textsc{Local}. The dialogue-macro results in Table~\ref{tab:write-main} favor editable two-stage writing over both alternatives on every ShareMem subset.

\begin{table}[!htb]
\centering
\begin{minipage}[t]{0.48\linewidth}
\centering
\caption{\textbf{Shared-pool redundancy.} Qwen, MemoryCode. E: editable two-stage; A: append-only. Ratio: entries per lexical cluster.}
\label{tab:write-redundancy}
{\small\setlength{\tabcolsep}{2.5pt}
\begin{tabular}{lrrrrrr}
\toprule
 & \multicolumn{2}{c}{Entries} & \multicolumn{2}{c}{Clusters} & \multicolumn{2}{c}{Ratio} \\
\cmidrule(lr){2-3}\cmidrule(lr){4-5}\cmidrule(lr){6-7}
Subset & E & A & E & A & E & A \\
\midrule
42 & 40 & 462 & 20 & 59 & 2.00 & 7.83 \\
23 & 17 & 467 & 9 & 64 & 1.89 & 7.30 \\
7 & 1 & 493 & 1 & 56 & 1.00 & 8.80 \\
\bottomrule
\end{tabular}
}
\end{minipage}\hfill
\begin{minipage}[t]{0.49\linewidth}
\centering
\caption{\textbf{MemoryCode induction cost.} Qwen, three subsets combined. Shared-stage counts; total induction includes both stages.}
\label{tab:write-cost}
{\small\setlength{\tabcolsep}{2.5pt}
\begin{tabular}{lrr}
\toprule
Quantity & Two-stage & Direct writing \\
\midrule
Manager calls & 304 & 2,530 \\
Input tokens & 694,832 & 4,070,841 \\
Output tokens & 81,139 & 367,143 \\
Total tokens & 775,971 & 4,437,984 \\
Total induction (M) & 3.76 & 7.43 \\
\bottomrule
\end{tabular}
\par}
\end{minipage}
\end{table}

For the lexical analysis in Table~\ref{tab:write-redundancy}, we remove words present in at least 90\% of a pool's entries and form connected components linking entries with word-set Jaccard similarity at least 0.5. Append-only enlarges the shared pool while increasing entries per cluster in every subset, indicating accumulation of lexically overlapping experiences. Figure~\ref{fig:write-pool-dynamics} tracks pool sizes against normalized source-session progress.

Table~\ref{tab:write-cost} quantifies the construction cost of the two write paths using construction-stage counts with the Qwen tokenizer. Processing accepted local edits instead of every raw session reduces shared-stage token use by a factor of 5.72 in this comparison. Including local induction, which costs approximately 2.99M tokens under either path, two-stage writing reduces total induction tokens by approximately 49\%. Together, the performance, pool, and cost comparisons support two-stage editable consolidation as an effective and economical way to maintain shared experiences.

\subsection{Experience retrieval and budget}
\label{app:budget}
We examine retrieval strategy and experience budget on Mind2Web. Each comparison fixes the experience pools and evaluation tasks. Scores are task-macro step success.

\paragraph{Retrieval strategy.}
Flat retrieval ranks accessible local and shared experiences together by similarity; scope-first retrieval additionally prioritizes matching scopes. Table~\ref{tab:flatpool} reports three-seed means at $K=5$: scope-first retrieval scores higher on Qwen, while the two strategies perform similarly on Gemma. Both shared configurations outperform Local on both backbones, distinguishing the benefit of shared access from the additional effect of scope priority.

\paragraph{Experience budget.}
Table~\ref{tab:scope-budget-details} varies the retrieval budget under both strategies. Same-site slots measure the fraction of retrieved entries associated with the evaluation website. The budget also bounds grounding demonstrations, so the sweep evaluates the combined guidance supplied at each budget.

\begin{table}[!htb]
\centering
\begin{minipage}[t]{0.48\linewidth}
\centering
\caption{\textbf{Scope priority and shared access.} Mind2Web; three-seed mean step success (\%), $K=5$. Scoped and Flat use ShareMem pools; $\Delta$ is Flat $-$ Scoped in pp.}
\label{tab:flatpool}
{\small\setlength{\tabcolsep}{3pt}
\begin{tabular}{lrrrr}
\toprule
Model & Local & Scoped & Flat & $\Delta$ \\
\midrule
Qwen & 43.06 & 47.16 & 46.22 & $-0.94$ \\
Gemma & 40.67 & 43.13 & 43.33 & $+0.20$ \\
\bottomrule
\end{tabular}
}
\end{minipage}\hfill
\begin{minipage}[t]{0.49\linewidth}
\centering
\caption{\textbf{Budget and retrieval relevance.} Qwen, Mind2Web, seed 42; frozen pools. Scores and slot fractions are percentages.}
\label{tab:scope-budget-details}
{\small\setlength{\tabcolsep}{3pt}
\begin{tabular}{rrrrr}
\toprule
 & \multicolumn{2}{c}{Step success} & \multicolumn{2}{c}{Same-site slots} \\
\cmidrule(lr){2-3}\cmidrule(lr){4-5}
$K$ & Scope-first & Flat & Scope-first & Flat \\
\midrule
1 & 41.45 & 40.57 & 100.00 & 86.53 \\
3 & 45.12 & 44.41 & 98.65 & 70.24 \\
5 & 47.48 & 46.76 & 94.87 & 58.95 \\
8 & 46.93 & 46.21 & 79.28 & 44.29 \\
10 & 47.21 & 47.18 & 68.29 & 38.36 \\
\bottomrule
\end{tabular}
}
\end{minipage}
\end{table}

Both strategies improve from one to five experiences and then plateau. In this Qwen seed-42 sweep, scope-first retrieval yields higher scores than flat retrieval at every tested budget, with gaps ranging from 0.03 to 0.88 percentage points. The separation is more pronounced in the proportion of same-site experiences retrieved, showing that scope priority concentrates guidance on the current website. These results support five entries as a compact operating point for this setting.

\subsection{Coverage and admission quality}
\label{app:gate}
\paragraph{Data assignment.}
We retain the same training demonstrations and evaluation tasks while changing their assignment to 20 users. Exclusive assigns each website's training tasks to one user and its evaluation tasks to another. Task-uniform assigns tasks to users through a seeded uniform mapping. Site-balanced distributes each website's tasks round-robin across users, with aligned user offsets for training and evaluation. Experience pools are rebuilt for each assignment and frozen during evaluation.

\begin{table}[!htb]
\centering
\caption{\textbf{Local coverage under different data assignments.} Qwen, Mind2Web; three-seed means over 252 evaluation tasks per seed. Coverage is the percentage of evaluation tasks with same-site local experience, measured from retrieval records. Scores are task-macro step success (\%); $\Delta$ is ShareMem $-$ Local in pp.}
\label{tab:assignment-coverage}
{\small
\begin{tabular}{lrrrrr}
\toprule
Data assignment & Coverage & None & Local & ShareMem & $\Delta$ \\
\midrule
Exclusive & 0.00 & 36.90 & 43.06 & 47.16 & $+4.10$ \\
Task-uniform & 49.60 & 36.66 & 43.98 & 45.28 & $+1.30$ \\
Site-balanced & 98.81 & 36.91 & 44.61 & 44.48 & $-0.13$ \\
\bottomrule
\end{tabular}
}
\end{table}

Table~\ref{tab:assignment-coverage} shows that the sharing gain shrinks as relevant experience becomes available locally. The same ordering holds in all three seeds. With nearly complete local coverage, the mean scores converge. Sharing is therefore most useful in this setting for supplying relevant experience to recipients who lack it; its incremental value diminishes when recipients already possess that experience.

\paragraph{Online admission.}
A separate sweep starts with empty experience pools and admits a completed rollout when its step-success rate reaches threshold $\tau$. We evaluate Qwen with 20 users and 252 tasks per seed over three seeds. Table~\ref{tab:gate} reports thresholds $\tau\geq0.5$ and combines performance with admitted-trajectory counts and final website coverage. Each condition generates its own online trajectories and updates its pools during execution.

\begin{table}[htbp]
\centering
\caption{\textbf{Online admission, performance, and coverage.} Qwen, Mind2Web; three-seed means. Scores are task-macro step success (\%). Website counts measure the union across all users' local pools for Local and the shared pool for ShareMem at the end of each run.}
\label{tab:gate}
{\small
\begin{tabular}{lrrrrr}
\toprule
Threshold $\tau$ & 0.50 & 0.67 & 0.75 & 0.90 & 0.999 \\
\midrule
\multicolumn{6}{c}{Task performance (\%)} \\
Local & 37.25 & 37.06 & 37.10 & 36.22 & 34.69 \\
\rowcolor{orange!10}ShareMem & 37.92 & 38.27 & 38.94 & 38.97 & 37.75 \\
\midrule
\multicolumn{6}{c}{Admitted trajectories} \\
Local & 184.00 & 77.00 & 67.33 & 27.33 & 23.00 \\
\rowcolor{orange!10}ShareMem & 192.33 & 88.33 & 79.67 & 43.33 & 34.33 \\
\midrule
\multicolumn{6}{c}{Websites represented in final pools} \\
Local & 57.33 & 36.67 & 30.33 & 9.67 & 7.33 \\
\rowcolor{orange!10}ShareMem & 58.00 & 40.67 & 36.33 & 23.00 & 15.00 \\
\bottomrule
\end{tabular}
}
\end{table}

Stricter admission reduces the number of contributing trajectories and narrows website coverage. Local performance declines as its experience supply contracts, whereas ShareMem maintains a narrower score range and retains broader shared-pool coverage at high thresholds. This online sweep captures the combined effects of admission, pool coverage, and subsequent agent interactions; the data-assignment experiment above isolates the allocation of a fixed demonstration set.

\subsection{Evidence source at fixed training-task coverage}
\label{app:mixed-source}
We retain all 1,009 training tasks and the exclusive data assignment. For each task, a nested random assignment selects a gold demonstration with probability $P$ and a cached no-memory rollout otherwise. Rollouts retain the agent's predicted actions, including successful and unsuccessful actions. Thus $P$ controls the evidence source. Both experience construction and attached demonstrations use the selected source, and the resulting pools are frozen for evaluation.

Table~\ref{tab:source-quality} complements the curves in Figure~\ref{fig:mixed-source}b. Replacing gold demonstrations with agent rollouts reduces ShareMem's advantage, with a larger endpoint decline for ShareMem than for Local. Local remains comparatively stable while at least half the training tasks use gold demonstrations. The contrast shows that useful cross-user transfer depends on the quality of the evidence from which experiences are constructed.

The retrieval audit for seeds 42 and 23 helps interpret this asymmetry. Under the exclusive assignment, Local supplies same-user experiences from other websites, whereas ShareMem provides site-matched shared experiences. Source quality therefore affects different guidance in the two configurations. The intervention jointly changes induced experience content and attached demonstrations; it measures sensitivity to the evidence source as a whole. Together with the online admission sweep, it motivates balancing evidence quality with the availability of relevant experience.

\subsection{Active preference retrieval}
\label{app:qwen-search-behavior}
We cross experience guidance with preference access on Qwen MemoryCode, using matched tasks and fixed preference stores and experience pools within each subset. Guidance is either disabled (\textsc{Pref}, with no local or shared experiences) or enabled (\methodname, with both local and shared experiences). All four configurations receive the same task-query top-10 initial preference recall. Initial-recall-only variants generate a response in one turn with retrieval-tool instructions removed; active variants can retrieve additional preferences during execution.

\begin{table}[!htb]
\centering
\begin{minipage}[t]{0.48\linewidth}
\centering
\caption{\textbf{Sensitivity to trajectory source.} Qwen, Mind2Web; three-seed mean step success (\%). $P$: gold sampling probability; $\Delta$: change from $P=1$ in pp.}
\label{tab:source-quality}
{\small\setlength{\tabcolsep}{3pt}
\begin{tabular}{rrrrr}
\toprule
 & \multicolumn{2}{c}{Step success} & \multicolumn{2}{c}{$\Delta$ from all-gold} \\
\cmidrule(lr){2-3}\cmidrule(lr){4-5}
$P$ & Local & ShareMem & Local & ShareMem \\
\midrule
1.00 & 41.92 & 46.60 & --- & --- \\
0.75 & 42.12 & 44.35 & $+0.20$ & $-2.25$ \\
0.50 & 41.68 & 44.12 & $-0.24$ & $-2.48$ \\
0.25 & 40.69 & 41.42 & $-1.22$ & $-5.17$ \\
0.00 & 40.39 & 40.70 & $-1.53$ & $-5.90$ \\
\bottomrule
\end{tabular}
}
\end{minipage}\hfill
\begin{minipage}[t]{0.49\linewidth}
\centering
\caption{\textbf{Experience--retrieval complementarity.} Qwen, MemoryCode; three-subset mean D-MS (\%). Off: preference-only access, with no experiences. On: both local and shared experience guidance. Gains are in pp.}
\label{tab:recall-three-seed}
{\small\setlength{\tabcolsep}{3pt}
\begin{tabular}{lrrr}
\toprule
Preference access & Off & On & Gain \\
\midrule
Initial recall only & 54.86 & 56.23 & $+1.37$ \\
Initial + active recall & 59.84 & 69.32 & $+9.48$ \\
\midrule
Active-retrieval gain & $+4.98$ & $+13.09$ & \\
\bottomrule
\end{tabular}
}
\end{minipage}
\end{table}

Table~\ref{tab:recall-three-seed} shows that experience guidance and active preference retrieval are complementary: each provides a larger gain when the other is available. The larger experience gain under active retrieval holds in all three subsets. This supports combining reusable guidance with access to user-specific constraints as execution progresses; the intervention changes both available information and the opportunity for further interaction.

\subsection{Cross-user preference interference}
\label{app:private-sharing}
\paragraph{Source labels and prompt instructions.}
\textsc{Share-all} explicitly labels preferences by source user and designates current-user preferences as authoritative, with other-user preferences serving only as weak hints. In VitaBench~2.0, the prompt explicitly prohibits copying other users' exact preference values into final actions. In MemoryCode, the prompt explicitly states that other-user rules must not override current-user rules. The audit examines misuse that persists despite these source labels and instructions.

\paragraph{Evidence and decision rule.}
We audit \textsc{Share-all} outputs from four backbones and three seeds. For failures exposed to other users' preferences, we assemble the user request, current-user preferences, source-labeled foreign preferences, final actions, and failed evaluation constraints. VitaBench~2.0 evidence includes tool-call arguments and responses; MemoryCode evidence includes generated code and the user's task-time coding rules. A positive judgment requires that the output adopts a concrete value traceable to an injected foreign preference, that the recipient's applicable preferences do not support that value, and that its use directly violates the failed constraint. Task-time rules are checked alongside stored preferences to distinguish foreign conventions from conventions already belonging to the recipient.

\paragraph{Screening and confirmation.}
Qwen3.6-27B screens the evidence at temperature zero. Candidate cases undergo provenance and current-user-rule checks, followed by three fresh judgments from the same model at temperature 0.7. Each repeat receives the evidence without the earlier verdict. Unanimous confirmation requires three valid positive judgments; majority confirmation requires at least two valid positive judgments among the three attempts.

\paragraph{Aggregation.}
For VitaBench~2.0, a failed subtask is confirmed when at least one of its failed constraints meets the selected voting threshold. MemoryCode uses failed tasks. We deduplicate within each run and sum counts across seeds. Table~\ref{tab:foreign-audit} reports the number of exposed failed tasks $N$, tasks containing a candidate record $F$, and confirmed tasks $C$. Each reported percentage is $100C/N$, including failures outside the candidate set in the denominator. Figure~\ref{fig:preference-interference} uses the unanimous column.
\begin{table}[htbp]
\centering
\caption{\textbf{Preference-interference screening and confirmation.} Share-all; three seeds pooled per backbone. $N$: exposed failed subtasks for VitaBench~2.0 and failed tasks for MemoryCode. $F$: tasks flagged for repeat review. Confirmation columns give task counts and percentages of $N$.}
\label{tab:foreign-audit}
{\small
\begin{tabular}{llrrrr}
\toprule
Benchmark & Backbone & $N$ & $F$ & Unanimous & Majority \\
\midrule
VitaBench~2.0 & Qwen & 1,557 & 47 & 38 (2.44\%) & 44 (2.83\%) \\
 & DeepSeek & 1,743 & 111 & 85 (4.88\%) & 97 (5.57\%) \\
 & Gemma & 1,785 & 26 & 21 (1.18\%) & 24 (1.34\%) \\
 & Gemini & 1,724 & 123 & 91 (5.28\%) & 109 (6.32\%) \\
 & Pooled & 6,809 & 307 & 235 (3.45\%) & 274 (4.02\%) \\
\midrule
MemoryCode & Qwen & 1,227 & 38 & 16 (1.30\%) & 24 (1.96\%) \\
 & DeepSeek & 1,224 & 12 & 4 (0.33\%) & 10 (0.82\%) \\
 & Gemma & 1,226 & 278 & 131 (10.69\%) & 210 (17.13\%) \\
 & Gemini & 1,227 & 46 & 15 (1.22\%) & 27 (2.20\%) \\
 & Pooled & 4,904 & 374 & 166 (3.38\%) & 271 (5.53\%) \\
\bottomrule
\end{tabular}
\par}
\end{table}

The cross-model pattern persists under both voting thresholds: Gemma has the highest confirmed misuse rate on MemoryCode, while Gemini and DeepSeek have the highest rates on VitaBench~2.0. The table also separates screened candidates from repeat-confirmed cases, particularly for MemoryCode. These rates describe model-confirmed misuse among exposed failures; screening misses remain unmeasured.

\paragraph{Case evidence.}
Table~\ref{tab:interference-cases} links recipient requirements, injected foreign preferences, and recorded outputs in four unanimously confirmed cases. The coding examples adopt another user's naming convention while failing the recipient's rule. The service examples apply another user's concrete choice to a request that leaves that choice unspecified, despite the recipient-specific evaluation requirement. These cases illustrate the audit's evidence criteria.

\begin{table}[!htbp]
\centering
\caption{\textbf{Examples of cross-user preference misuse.} Share-all, seed 23; two cases per benchmark, each confirmed by three positive repeat judgments. Foreign entries were source-labeled and presented as non-authoritative hints. Preference text and requests are condensed. Code identifiers are preserved. Requirements come from task-time rules or evaluation constraints; outputs are recorded outputs.}
\label{tab:interference-cases}
{\small
\setlength{\tabcolsep}{4pt}
\begin{tabular}{@{}>{\raggedright\arraybackslash}p{.14\linewidth}>{\raggedright\arraybackslash}p{.26\linewidth}>{\raggedright\arraybackslash}p{.23\linewidth}>{\raggedright\arraybackslash}p{\dimexpr.37\linewidth-24pt\relax}@{}}
\toprule
Case & Task and recipient requirement & Injected foreign preference & Observed output and violation \\
\midrule
MemoryCode\newline DeepSeek
& Implement greatest common divisor. Task-time function names must end in \texttt{\_y}.
& Include \texttt{chx} in every function name.
& Generates \texttt{fn\_gcd\_chx}. The name adopts \texttt{chx} and fails the required \texttt{\_y} suffix. \\
\midrule
MemoryCode\newline Gemma
& Validate a binary search tree. Function arguments must start with \texttt{g\_}.
& End function argument names with \texttt{\_j}.
& Uses \texttt{root\_j}, \texttt{node\_j}, \texttt{low\_j}, and \texttt{high\_j}. These arguments adopt the foreign suffix and omit the required prefix. \\
\midrule
VitaBench~2.0\newline Qwen
& Order a named milk tea. Current-user memory specifies extra ice; the evaluation requires that option.
& Another user previously ordered the same drink hot.
& Creates an order with temperature set to hot and attributes this choice to the recipient's prior preference. The extra-ice constraint fails. \\
\midrule
VitaBench~2.0\newline DeepSeek
& Book round-trip flights to Seoul. The request specifies dates; the evaluation requires economy class.
& Another user prefers business-class flights.
& Books and pays for an outbound business-class ticket, as reported in its final itinerary. The economy-class constraint fails. \\
\bottomrule
\end{tabular}
}
\end{table}

\clearpage
\section{Prompts}
\label{app:prompts}

We present key VitaBench~2.0 prompts for memory construction and personalized execution.

\subsection{Local experience updates}
\label{app:prompt-local}
\label{tab:prompt-update}

\Needspace{6\baselineskip}
\begin{memoryprompt}{paperblue}{Local experience update | System prompt}
You maintain a local experience pool for one user and one service domain. The pool stores reusable procedures learned from this user's successful trajectories, but never stores the user's concrete preference values. Each entry should have one recognizable task trigger and should tell the executor which preference dimensions may matter, how to resolve missing aspects, how to act, and how to verify or recover. The candidate is evidence from a successful trajectory and supporting historical interactions. Output only JSON operations.
Granularity and coverage policy:
- Merge local experiences when they have the same primary task trigger, preference-dimension needs, decision logic, and tool procedure, even if their source entities differ.
- Keep experiences separate when the task trigger, required aspects, tool dependencies, decision stage, or recovery behavior differs.
- Enrich an existing same-trigger experience when new successful evidence reveals a reusable missing preference dimension or safer branch.
- Never create an entire-domain umbrella experience and never preserve a concrete preference merely because it recurs in this user's history.
Preference-dimension and control-flow policy:
- Treat historical interactions only as supporting evidence for abstract preference dimensions; derive procedural claims primarily from the successful trajectory.
- Auto-recalled preferences are coarse and potentially incomplete. Preserve focused search for unresolved independent aspects rather than one generic retrieval step.
- Preserve useful conditional branches. Permit a backward return only for a specific unresolved field or failed operation, only with a materially narrower query or changed action, and at most once for that field or operation.
- Do not encode unnecessary confirmation, waiting, or exhaustive comparison. Current task data and observations override the experience.
Allowed operations:
- add: add an experience not covered by related entries
- replace: consolidate the same task trigger or make its preference-dimension coverage, decisions, or bounded recovery materially safer or more complete; content must be the complete replacement entry
- remove: remove one related entry only when the candidate proves it wrong or unsafe
Never store or infer user-specific preference values, ids, names, addresses, dates, or prices; use {variable} placeholders. Do not write concrete examples or e.g. clauses. Prefer replace over add only when the candidate has the same primary task trigger and substantially overlapping preference-dimension and procedural logic.
Manager decisions are uniform across benchmarks:
- add: a reusable trigger/procedure is not covered by related entries.
- replace: the same trigger is covered, but the candidate is safer or more complete; return one complete replacement.
- remove: an existing entry is proven wrong or unsafe.
- Before add/replace, rewrite every source-user or source-task value as a semantic {slot}; never copy examples.
- no-op: the candidate is already covered, one-off, malformed, or has no reusable procedure after abstraction.
\end{memoryprompt}

\Needspace{6\baselineskip}
\begin{memoryprompt}{paperblue}{Local experience update | User-message template}
Related local experiences from the same user and service domain:
<retrieved nearby entries>
Candidate experience from a successful trajectory:
<candidate experience>
First compare primary task trigger, applicable preference dimensions, tool dependencies, decision logic, and recovery branches. Add a genuinely distinct reusable experience; replace a same-trigger entry when the candidate improves abstract aspect coverage or bounded control flow; otherwise no-op. Ensure a complete replacement contains no source values and does not turn distinct task families into an umbrella experience.
Return strict JSON only:
{"operations":[{"action":"add","content":"..."},{"action":"replace","old_text":"unique substring from old entry","content":"..."},{"action":"remove","old_text":"unique substring from old entry"}]}
For replace/remove, old_text must be a unique substring copied from a related entry. For replace, content must be the complete replacement experience, including the title and all retained steps. If the pool should not change, return {"operations": []}.
\end{memoryprompt}

\clearpage
\subsection{Shared experience updates}
\label{app:prompt-shared}
\label{fig:prompt-shared}

\Needspace{6\baselineskip}
\begin{memoryprompt}{paperteal}{Shared experience update | System prompt}
You maintain a retrieval-oriented shared pool of reusable service-task experiences. Each entry should have one recognizable task trigger and a Procedure that helps an executor identify applicable preference dimensions, resolve missing aspects with focused searches, act, verify, and recover safely. The candidate is cross-user evidence, not a direct command. Output only JSON operations.
Granularity and coverage policy:
- Maintain a compact, diverse set of independently retrievable experiences for distinct task families or decision mechanisms.
- Merge experiences that solve the same underlying problem and differ only in products, merchants, locations, wording, or source users.
- Keep experiences separate when their primary task trigger, required preference dimensions, tool dependencies, decision stage, or failure recovery differs.
- Shared meta-steps such as retrieve preferences, search, and verify do not by themselves make different experiences duplicates.
- Never create or expand a universal experience that attempts to cover an entire service domain.
Preference-dimension and control-flow policy:
- An experience should name the required and conditional preference dimensions generally needed for its task family, without storing their values.
- Auto-recalled preferences are coarse and potentially incomplete. Preserve guidance to search separately for unresolved independent aspects.
- Preserve useful conditional branches. Permit a backward return only for a specific unresolved field or failed operation, only with a materially narrower query or changed action, and at most once for that field or operation.
- Remove unnecessary option presentation, waiting, exhaustive comparison, and confirmation when the task already authorizes execution.
- Current instructions, current-user preferences, profile, context, and tool observations must override experience guidance.
Allowed operations:
- add: add an experience not covered by related entries
- replace: consolidate the same task trigger or make its preference-dimension coverage, decisions, or bounded recovery materially safer or more complete; content must be the complete replacement entry
- remove: remove one related entry only when the candidate proves it wrong or unsafe
Never store or infer user-specific preference values, ids, names, addresses, dates, or prices; use {variable} placeholders. Do not write concrete examples or e.g. clauses.
\end{memoryprompt}

\Needspace{6\baselineskip}
\begin{memoryprompt}{paperteal}{Shared experience update | User-message template}
Related shared experiences:
<retrieved nearby entries>
Candidate experience from a successful trajectory:
<candidate experience>
Before deciding, compare primary task trigger, required and conditional preference dimensions, tool dependencies, decision stage, and recovery branches. Merge entity-level variants of the same underlying experience, but preserve independently applicable task families or decision mechanisms. Use the candidate to improve abstract preference-dimension coverage and bounded control flow while removing every source-user value. Do not create a domain-wide umbrella experience, an unbounded retry loop, or a mandatory confirmation step.
Return strict JSON only:
{"operations":[{"action":"add","content":"..."},{"action":"replace","old_text":"unique substring from old entry","content":"..."},{"action":"remove","old_text":"unique substring from old entry"}]}
For replace/remove, old_text must be a unique substring copied from a related entry. If the pool should not change, return {"operations": []}.
\end{memoryprompt}

\clearpage
\subsection{Experience extraction}
\label{app:prompt-extraction}
\label{tab:prompt-execution}

\Needspace{6\baselineskip}
\begin{memoryprompt}{PromptShared}{Experience extraction | System prompt}
You extract retrieval-oriented reusable experiences from successful service trajectories.
Use this experience representation:
## <snake_case_name>
Scope: domain=<delivery|instore|ota>
Applies when: <transferable task trigger without source values>
Inputs: {slot_1}, {slot_2}
Procedure:
[step 1] <imperative action, check, or conditional branch>
[step 2] <imperative action, check, or conditional branch>
[step 3] <verification or bounded recovery step>
Caution: <failure mode and boundary>
Evidence roles:
- The successful instruction and trajectory are the primary evidence for what procedure worked, including tool order, decisions, and recovery behavior.
- Historical user interactions are supporting evidence for identifying which preference dimensions can matter to this task family. Do not imitate their concrete preference values, entities, incidental dialogue, or confirmation style.
- Abstract every task-specific or user-specific value into a semantic `{slot}` before returning the experience.
Extraction rules:
- Produce one reusable experience with 3-8 steps, or an empty string when no reusable procedure exists.
- Give the experience one recognizable task trigger. Do not create a universal experience for an entire service domain.
- Make the preference-retrieval dependency operational. Early in Procedure, identify the required and conditionally applicable preference dimensions for this task family, such as merchant or item preference, customization, context-specific address, time, accessibility, health, allergy, or safety constraints.
- Tell the executor to classify applicable aspects as resolved or unresolved using the current request, current-user preferences, profile, context, and tool observations. Search preferences separately for unresolved independent aspects instead of issuing one broad query.
- Auto-recalled preferences provide only coarse initial evidence and may omit relevant aspects. Their presence does not imply that every personalized field is resolved.
- Preserve essential service/tool ordering, decision rules, success checks, and useful recovery behavior from the successful trajectory.
- Conditional branches are allowed. A backward return is allowed only for a specific unresolved field or failed operation, only when a narrower query or materially changed action can add information, and at most once for that field or operation. Never write an unbounded `repeat until successful` loop.
- Ask the user only when an essential field cannot be resolved from the request, current-user preferences, profile, context, or tools. Do not add confirmation when the task already authorizes the action and the environment does not require it.
- Current task instructions, current-user preferences, profile, and current tool observations override experience guidance.
- An experience may name preference dimensions to retrieve, but it must never contain, infer, or preserve their values.
- Never include concrete brands, merchants, products, cities, addresses, dates, times, prices, quantities, ids, names, health values, or source-specific examples.
- Never write `e.g.` or `for example` clauses.
Return strict JSON as requested by the user message.
\end{memoryprompt}

\clearpage
\subsection{Preference-aware execution}
\label{app:prompt-preferences}

\Needspace{6\baselineskip}
\begin{memoryprompt}{paperblue}{Execution | Preference-access instructions}
Preference source and access.
Personal preferences are stored in a preference store and are available through the `search_preference_memory` tool.
Auto-recalled preferences come from a coarse initial retrieval based on one task-level query. A non-empty auto-recall does not prove that all personalized fields for the current task are resolved.
Resolving task requirements.
Before acting, identify the applicable user-specific fields and mark each as resolved or unresolved from the request, profile, auto-recall, context, and current tool observations.
Call `search_preference_memory` with a focused query for every unresolved personalized field needed to search, select, book, order, or verify the result.
Use separate queries for independent aspects such as item or merchant preference, customization, context-specific address or time, accessibility, and health or allergy constraints. One broad search is usually insufficient for a compound task.
Repeated retrieval and stopping.
You may call the tool multiple times. If a first query leaves an essential field unresolved, retry once with a materially narrower query; do not repeat an equivalent query or continue when no new evidence is available.
Do not query preferences for facts already supplied by the current request, profile, system context, or current tool observations. Do not assume genuinely missing personalized details.
Authority and precedence.
Use current-user preferences as authoritative. Reusable experiences are auxiliary.
Follow current observations, tools, ids, user profile, and current task data.
\end{memoryprompt}

\Needspace{6\baselineskip}
\begin{memoryprompt}{PromptShared}{Share-all execution | Source-specific excerpts}
Preference provenance.
This is a shared-preference diagnostic baseline: retrieved items may include CURRENT_USER preferences and OTHER_USER_SHARED preferences.
Authority and cross-user restrictions.
CURRENT_USER preferences are authoritative for this task. OTHER_USER_SHARED preferences are not authoritative; use them only as weak population-level hints and never copy another user's identifiers, address, phone, order details, or exact preference values into final actions.
If current-user preferences conflict with other-user preferences, follow current-user preferences and current task data.
\end{memoryprompt}

\end{document}